%% file: neurips_2026.tex
\pdfoutput=1
\documentclass{article}

\usepackage{neurips_2026}

\usepackage[T1]{fontenc}
\usepackage{hyperref}
\usepackage{booktabs}
\usepackage{amsfonts}
\usepackage{amsmath}
\usepackage{microtype}
\usepackage{xcolor}
\usepackage{colortbl}
\usepackage{graphicx}
\usepackage{multirow}
\usepackage{titletoc}
\usepackage[capitalize,noabbrev]{cleveref}
\usepackage{enumitem}

\newcommand{\PV}{\ensuremath{\mathrm{PV}}}
\newcommand{\PVFormer}{\textsc{PVFormer}}
\newcommand{\PVNN}{\textsc{PVNN}}
\newcommand{\Hypformer}{{Hypformer}}
\newcommand{\darkred}[1]{{\color{red!50!black} #1}}
\newtheorem{theorem}{Theorem}
\hypersetup{hidelinks}
\title{PVFormer: Proper Velocity Transformer for Stable and Scalable Hyperbolic Representation Learning}

\author{%
  Xianglong Shi$^{1}$ \quad
  Nicu Sebe$^{2}$ \quad
  Bernhard Sch{\"o}lkopf$^{3}$ \quad
  Ziheng Chen$^{3,2}$\thanks{Corresponding author.} \\
  {\normalfont $^{1}$University of Science and Technology of China} \\
  {\normalfont $^{2}$University of Trento} \\
  {\normalfont $^{3}$Max Planck Institute for Intelligent Systems, Tübingen} \\
  {\normalfont\texttt{xlshi@mail.ustc.edu.cn, ziheng\_ch@163.com}}
}

\begin{document}

\maketitle

\begin{abstract}
Hyperbolic representations have shown remarkable success on hierarchical and tree-like data, and Transformers have become a central architecture for representation learning. However, a principled construction of hyperbolic Transformers remains challenging because the self-attention layer decomposes into three coupled primitives, transformation, similarity, and aggregation, that should all respect the underlying geometry. Existing hyperbolic Transformer designs are typically built in the Poincar\'e or Lorentz models, where bounded domains, time-like constraints, tangent-space detours, or projection steps make it difficult to keep the full block intrinsic and scalable. To address this gap, we propose \PVFormer{}, to the best of our knowledge, the first Transformer architecture that operates natively in proper velocity (\PV) space. The proper velocity model provides an unconstrained representation of hyperbolic geometry, and \PVFormer{} exploits this structure to redesign the Transformer block in \PV{} space. On the transformation side, we use \PV{} homomorphism transformations to update query and value representations. On the similarity side, we generalize Euclidean inner-product attention with a \PV{} Busemann score inspired by Busemann-based hyperbolic learning and a \PV{} point-to-hyperplane key transformation. On the aggregation side, we replace Euclidean averaging with \PV{} midpoint aggregation and further derive a scalable \PV{} linear attention variant for large graphs. Extensive experiments support the effectiveness of \PVFormer{} across graph, text, and vision benchmarks.
\end{abstract}
\section{Introduction}
Hyperbolic geometry provides a natural representation for hierarchical and tree-like data because negative curvature offers exponential representation capacity \citep{nickel2017poincar,nickel2018learning,ganea2018hyperbolic,liu2019hyperbolic,chami2019hyperbolic}. This property has driven successful applications in graphs~\citep{gcn2017, yao2018graph,yang2024hypformer}, visual categories~\citep{bdeir2023fully,shi2026intrinsic}, and biological sequences \citep{shi2026intrinsic,chen2026proper}, and recent work has moved from hyperbolic embeddings toward hyperbolic neural networks that operate within hyperbolic space. However, as emphasized by~\PVNN{}~\citep{chen2026proper}, the choice of hyperbolic model is central to neural architecture design \citep{shi2026intrinsic,chen2026proper}. Poincar\'e and Lorentz models provide powerful formulations and many closed-form operations, but bounded domains, explicit manifold constraints, tangent-space detours, and projection steps can introduce numerical and implementation difficulties. In particular, Poincar\'e embeddings that approach the boundary can make computations unstable and cause gradients to vanish~\citep{chen2026proper}.

These issues become especially important for Transformers, whose self-attention layer is a coupled pipeline rather than a single similarity function. Each block must transform hidden states into query, key, and value representations, compare queries and keys, and aggregate values \citep{vaswani2017attention}. Existing hyperbolic attention and Transformer methods make important progress toward this goal, but they often align only part of the pipeline with hyperbolic geometry \citep{gulcehre2018hyperbolic,yang2024hypformer}. Euclidean or tangent-space transformations followed by restoration steps preserve feasibility but weaken the link between learned updates and intrinsic geometry \citep{chami2019hyperbolic,yang2024hypformer}; distance-based similarities encode proximity but can discard directionality, while tangent-space inner products depend on a base point \citep{gulcehre2018hyperbolic,yang2024hypformer}; and log/exp-map or projection-based aggregation can add computational and numerical burden \citep{park2024hyperbolic}. The challenge is therefore not merely to replace the Euclidean dot product, but to realize transformation, similarity, and aggregation in one coherent hyperbolic computational model.

Proper velocity (\PV) space suggests such a route. It represents the same hyperbolic geometry as the Poincar\'e and Lorentz models using unconstrained vectors, while retaining Riemannian operations for hyperbolic learning through the \PV{} metric and gyrovector structure \citep{ungar2022analytic,chen2026proper}. This model avoids bounded-ball management and explicit hyperboloid restoration, making it attractive for repeatedly composed neural blocks. The \PVNN{} demonstrates that \PV{} space can support stable hyperbolic neural layers, but it does not instantiate the attention pipeline required by Transformers. A native proper velocity Transformer must still define geometry-aware transformations, directional similarity, intrinsic aggregation, and a scalable attention path within the same \PV{} representation.

Motivated by this gap, we propose \PVFormer{}, to the best of our knowledge, the first Transformer architecture operating natively in proper velocity space. \PVFormer{} keeps the attention pipeline in \PV{} space by using \PV{} homomorphism transformations for query and value updates, a \PV{} point-to-hyperplane transformation that produces direction-like keys, \PV{} Busemann attention for curvature-aware similarity \citep{atigh2021hyperbolic}, and \PV{} midpoint aggregation in place of Euclidean weighted sums. For large graphs, we further use a \PV{} linear attention path that avoids materializing a quadratic Busemann score matrix while preserving the midpoint-based aggregation principle. We evaluate \PVFormer{} across graph, text, and vision benchmarks under a unified protocol, showing that proper velocity space can serve not only as a stable hyperbolic representation model but also as the computational foundation for Transformer attention.

Our contributions are summarized as follows:
\begin{itemize}[leftmargin=7pt, itemsep=0.2em, topsep=0.2em]
    \item We formulate the three primitives of attention, transformation, similarity, and aggregation, directly in \PV{} space through \PV{} homomorphism transformations, \PV{} point-to-hyperplane keys for Busemann attention, and \PV{} midpoint aggregation.
    \item We derive \PV{} linear attention, a scalable linear-attention variant that avoids materializing a quadratic Busemann score matrix while preserving \PV{} midpoint aggregation for large graph settings.
    \item We develop \PVFormer{}, to the best of our knowledge, the first Transformer architecture operating natively in proper velocity space and the first hyperbolic Transformer to generalize dot-product attention through the Busemann function, which recovers the Euclidean dot product as $K\to0^-$.
    \item We show that \PVFormer{} achieves stronger predictive performance and faster attention computation across graph, text, and vision benchmarks under a unified experimental protocol.
\end{itemize}
\vspace{-1mm}
\section{Related Work}
\subsection{Hyperbolic neural networks and geometric models}
Hyperbolic representation learning exploits negative curvature for hierarchical or tree-like structure, beginning with Poincar\'e and Lorentz embeddings and extending to neural and graph architectures through gyrovector and Riemannian operations \citep{cannon1997hyperbolic,nickel2017poincar,nickel2018learning,ganea2018hyperbolic,liu2019hyperbolic,chami2019hyperbolic,bronstein2016geometric,bronstein2021geometric}. Model choice directly affects implementation: Poincar\'e models are bounded, Lorentz models introduce an explicit time-like dimension, and both can exhibit numerical sensitivity in composed neural layers \citep{mishne2022numerical,yu2021representing}. Fully Lorentz and intrinsic Lorentz networks keep operations on the hyperboloid \citep{chen2021fully,klis2026fast,shi2026intrinsic}, whereas PVNN~\citep{chen2026proper} develops closed-form layers in an unconstrained \PV{} model derived from gyrovector structures \citep{ungar2022analytic,chen2026proper}. \PVFormer{} builds on this model-level opportunity but targets the Transformer block itself, rather than using \PV{} only as a layer family or representation conversion.

\subsection{Hyperbolic attention, Busemann functions, and Transformers}
Transformers organize representation learning around transformations, similarity, and aggregation \citep{vaswani2017attention}; graph Transformers and efficient graph attention variants further emphasize the scalability constraints of this decomposition \citep{ying2021transformers,rampek2022recipe,shirzad2023exphormer,wu2023nodeformer,wu2023simplifying}. Hyperbolic Attention Networks and \Hypformer{} move attention or complete Transformer modules into hyperbolic geometry, with \Hypformer{} providing the main Lorentz-model linear-attention reference for our experiments \citep{gulcehre2018hyperbolic,yang2024hypformer}. In parallel, Busemann-based methods use ideal-boundary potentials for hyperbolic similarity and classification, and hyperbolic vision Transformers illustrate broader non-Euclidean Transformer designs \citep{atigh2021hyperbolic,alyoussef2026hexformer}. \PVFormer{} differs by placing the transformation--similarity--aggregation pipeline in \PV{} space, combining homomorphic updates, \PV{} Busemann attention, \PV{} midpoint aggregation, and \PV{} linear attention in one native \PV{} Transformer.

\section{Preliminaries}

\subsection{Riemannian geometry}
\begin{figure}[t]
\centering
\includegraphics[width=0.85\linewidth]{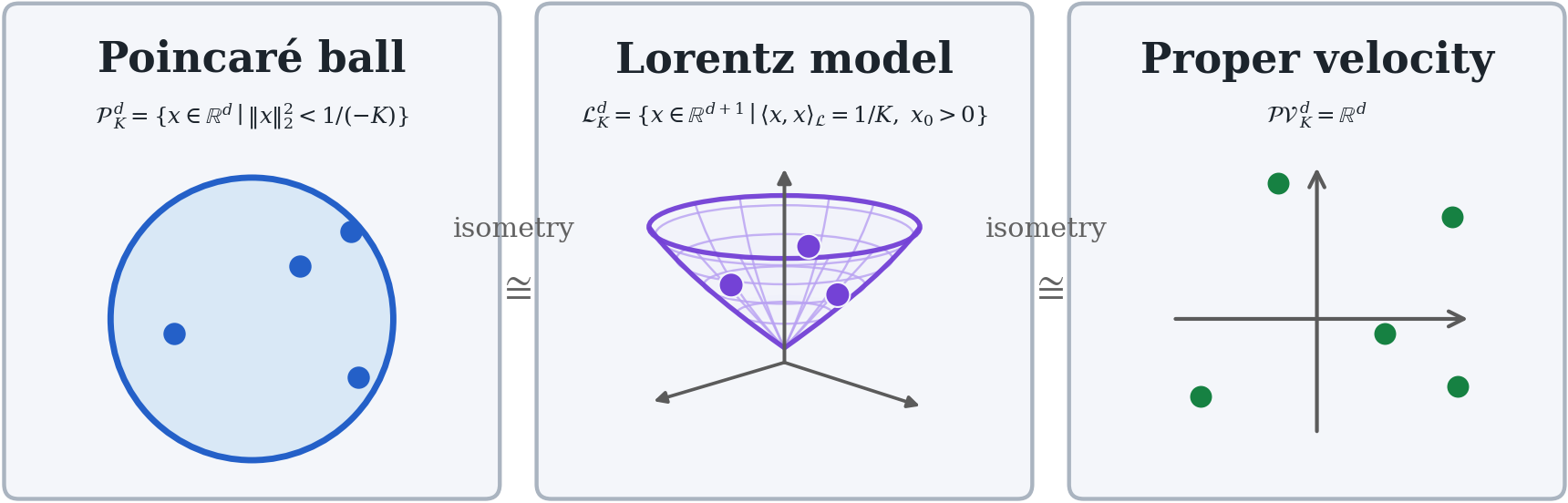}
\caption{Comparison of three two-dimensional hyperbolic models. \PV{} represents the same hyperbolic geometry as the Poincar\'e and Lorentz models, but removes equality or inequality constraints.}
\label{fig:geometry}
\end{figure}

We use standard Riemannian notation: $(\mathcal{M},g)$ denotes a smooth manifold with tangent inner product $g_x$ and norm $\|v\|_x=\sqrt{g_x(v,v)}$; $d$, $\operatorname{Exp}_x$, $\operatorname{Log}_x$, and $\operatorname{PT}_{x\to y}$ denote geodesic distance, exponential map, logarithmic map, and parallel transport \citep{lee2018riemannian}. An isometry is a smooth bijection whose differential preserves the tangent inner products. Hyperbolic space is the simply connected manifold of constant negative curvature \citep{cannon1997hyperbolic}. As shown in \Cref{fig:geometry}, the Poincar\'e, Lorentz, and proper velocity models describe this same geometry with different constraints: a bounded ball, an ambient hyperboloid, or an unconstrained \PV{} representation connected to gyrovector constructions \citep{ungar2022analytic,chen2026proper}.

\subsection{Proper velocity space}
Let $K<0$ and let $\mathcal{PV}^d_K=\mathbb{R}^d$ be the $d$-dimensional proper velocity model. Points are stored as unconstrained vectors, but tangent vectors are measured by the \PV{} metric~\citep{chen2026proper}:
\begin{equation}
\beta_u=\frac{1}{\sqrt{1-K\|u\|^2}},\qquad
g_u(a,b)=\langle a,b\rangle + K\beta_u^2\langle u,a\rangle\langle u,b\rangle,
\label{eq:pv_metric}
\end{equation}
for $a,b\in T_u\mathcal{PV}^d_K\simeq\mathbb{R}^d$. Because $K<0$, $\beta_u$ is finite for every $u\in\mathbb{R}^d$, unlike the bounded Poincar\'e ball or the Lorentz hyperboloid constraint. \PV{} space also has gyroaddition $\oplus_U$ and scalar gyromultiplication $\otimes_U$ that generalize Euclidean vector operations \citep{ungar2022analytic,chen2026proper}; \PVFormer{} uses this structure to keep transformations, similarity, and aggregation \PV{}-compatible.
\PV{} is not a different geometry from the Lorentz model but an unconstrained model of the same hyperbolic manifold~\citep{chen2026proper}. The lift $\iota_{\PV\to\mathbb{L}}(x)=(\sqrt{\|x\|^2-1/K},x)$ maps a \PV{} point to the Lorentz hyperboloid, so the \PV{} layers used by \PVFormer{} can be viewed as Lorentz-model hyperbolic operations represented in \PV{} space, with the time-like constraint absorbed by the metric and closed-form \PV{} operations. This relation lets \PVFormer{} retain the geometric meaning of Lorentz-based hyperbolic Transformers while avoiding explicit hyperboloid projection or restoration inside the block.

\subsection{Busemann functions}
For a proper geodesic metric space $(X,d)$ and a unit-speed geodesic ray $\gamma:[0,\infty)\to X$, the Busemann function is
\begin{equation}
B^{\gamma}(x)=\lim_{t\to\infty}\bigl(d(x,\gamma(t))-t\bigr),
\end{equation}
where $x\in X$ \citep{bridson2013metric}. In Euclidean space, the ray $\gamma(t)=tv$ with $\|v\|=1$ yields $B^{v}(x)=-\langle x,v\rangle$.
Thus, Busemann potentials recover directional inner products in the flat case and provide the similarity primitive used below for curvature-aware attention.

\section{Method}
In this section, we present \PVFormer{}. Following the three-part view of attention, we redesign the Transformer block directly in proper velocity space. First, we construct \emph{query} ($Q$) and \emph{value} ($V$) updates as \PV{} gyro-homomorphic transformations induced by origin-preserving orthogonal maps. These updates are Riemannian isometries of the PV metric and therefore play the role of geometry-preserving analogues of Euclidean linear projections. Second, we reformulate dot-product similarity through the Busemann function and obtain a curvature-aware \PV{} Busemann attention score. Third, because Busemann scoring requires direction-like keys, we introduce a \PV{} point-to-hyperplane key transformation that maps hidden states to the origin tangent space. Finally, for large graphs, we use \PV{} linear attention, which replaces quadratic Busemann scoring with kernelized linear weights while preserving \PV{} midpoint aggregation. The resulting pipeline keeps transformation, similarity, and aggregation in the same \PV{} computational model instead of alternating between Euclidean updates and post-hoc geometric restoration.

\subsection{PV homomorphism transformation}
\label{subsec:pv_homomorphism}

In Euclidean attention, the $Q$ and $V$ updates are linear homomorphisms: they preserve vector addition and scalar multiplication. We generalize this role to proper velocity space by requiring the corresponding updates to preserve PV gyroaddition and scalar gyromultiplication. Let $K<0$ and let $\mathcal{PV}^d_K=\mathbb{R}^d$ be equipped with the PV metric in \Cref{eq:pv_metric}. For $x\in\mathcal{PV}^d_K$, write $\beta_x=\frac{1}{\sqrt{1-K\|x\|^2}}$.
The PV gyroaddition and scalar gyromultiplication can be written as \citep{chen2026proper,ungar2022analytic}
\begin{align}
x\oplus_U y
&=x+y+
\left(
\frac{1-\beta_y}{\beta_y}
-K\frac{\beta_x}{1+\beta_x}\langle x,y\rangle
\right)x,
\\
t\otimes_U x
&=\frac{\sinh\bigl(t\operatorname{arsinh}(\sqrt{-K}\|x\|)\bigr)}{\sqrt{-K}\|x\|}x,
\qquad t\otimes_U 0=0.
\end{align}

\begin{theorem}[PV isotropy homomorphism]
\label{thm:pv_isotropy_homomorphism}
Let $R\in SO(d)$ and define
\begin{equation}
\Phi_R^{\PV}:\mathcal{PV}^d_K\to\mathcal{PV}^d_K,
\qquad
\Phi_R^{\PV}(x)=Rx.
\end{equation}
Then $\Phi_R^{\PV}$ is both a gyrovector homomorphism and a Riemannian isometry:
\begin{align}
\Phi_R^{\PV}(x\oplus_U y)&=\Phi_R^{\PV}(x)\oplus_U\Phi_R^{\PV}(y),
\\
\Phi_R^{\PV}(t\otimes_U x)&=t\otimes_U\Phi_R^{\PV}(x),
\\
g_{Rx}(Ru,Rv)&=g_x(u,v).
\end{align}
\end{theorem}
Although $\mathcal{PV}^d_K$ is stored as an unconstrained array, \Cref{thm:pv_isotropy_homomorphism} shows that $R$ is not being used as an arbitrary Euclidean layer: it preserves the PV algebraic operations and the PV Riemannian metric. We provide the proof in \Cref{app:proof_isotropy} and use the theorem here as a design principle for constraining \PVFormer{} query and value updates to geometry-preserving rotations rather than arbitrary Euclidean projections.

This construction specializes the \PV--Lorentz relation above to homomorphism transformations. Conjugating the Lorentz isotropy map $\operatorname{diag}(1,R)$ through the \PV--hyperboloid isometry gives exactly $x\mapsto Rx$ in the \PV{} model. Therefore, the origin-preserving Lorentz homomorphism used for hyperbolic $Q/V$ updates becomes an ordinary orthogonal rotation in \PV{} space, while retaining the same geometric meaning.

No bias term is added: a biased map would not preserve the gyro identity $0$, and an unconstrained linear map would not generally preserve the PV metric. The key representation is handled separately by the \PV{} point-to-hyperplane key transformation in \Cref{subsec:pv_pt2hyperplane_key}, because Busemann similarity requires direction-like keys rather than another point-valued homomorphic update. Our construction uses equal-dimensional rotations in $SO(d)$. Its expressiveness and possible dimension-changing extensions are discussed in \Cref{app:homomorphic_discussion}.

\subsection{PV Busemann attention}
\label{subsec:pv_busemann_attention}

\PVFormer{} defines similarity by rewriting Euclidean dot-product attention as a Busemann potential. For nonzero $k$ and $u_k=k/\|k\|$, the identity $\langle q,k\rangle=-\|k\|B^{u_k}(q)$ expresses the dot product as a scaled directional Busemann function. Pulling the Lorentz Busemann function back through the \PV--hyperboloid isometry gives the following score entirely in \PV{} space.

\begin{theorem}[PV Busemann function]
\label{thm:pv_busemann_function}
Let $K<0$ and $q\in\mathcal{PV}^d_K$. For a unit direction $u\in T_0\mathcal{PV}^d_K\simeq\mathbb{R}^d$ with $\|u\|=1$, let $\gamma_u(t)=\operatorname{Exp}_0(tu)$ be the unit-speed geodesic ray from the origin. The \PV{} Busemann function associated with this ray is
\begin{equation}
B^{u}_{\PV,K}(q)
=\lim_{t\to\infty}\bigl(d_{\PV}(q,\gamma_u(t))-t\bigr)
=\frac{1}{\sqrt{-K}}\log\left(\sqrt{1-K\|q\|^2}-\sqrt{-K}\langle q,u\rangle\right).
\end{equation}
For any nonzero tangent key $k\in T_0\mathcal{PV}^d_K\setminus\{0\}$, the corresponding \PV{} Busemann similarity is
\begin{equation}
s_{\PV}(q,k)
=-\|k\|B^{k/\|k\|}_{\PV,K}(q)
=-\frac{\|k\|}{\sqrt{-K}}
\log\left(
\sqrt{1-K\|q\|^2}
-\sqrt{-K}\left\langle q,\frac{k}{\|k\|}\right\rangle
\right).
\label{eq:pv_busemann_similarity_closed}
\end{equation}
Moreover, $s_{\PV}(q,k)\to\langle q,k\rangle$ as $K\to0^-$.
\end{theorem}

The derivation, positivity of the logarithm, and flat-limit proof are given in \Cref{app:proof_pv_busemann}.

\textbf{PV Busemann logits.}
For each head, the query is $q_i=\Phi_{R_Q}^{\PV}(u_i)$ and the point-to-hyperplane key module in \Cref{subsec:pv_pt2hyperplane_key} returns $k_j$. We use $\hat{k}_j=\frac{k_j}{\|k_j\|+\varepsilon}$ and $\rho_i=\sqrt{1-K\|q_i\|^2}$, and compute the implemented logit as
\begin{equation}
S_{ij}^{\PV}
=-\frac{\tau_h}{\sqrt{-K}}
\log\left(\rho_i-\sqrt{-K}\langle q_i,\hat{k}_j\rangle+\varepsilon\right),
\label{eq:pv_busemann_score}
\end{equation}
where $\tau_h>0$ is learnable; replacing it by the key norm recovers the strict dot-product generalization in \Cref{thm:pv_busemann_function}.

\begin{theorem}[PV midpoint aggregation]
\label{thm:pv_midpoint}
Let $v_1,\ldots,v_N\in\mathcal{PV}^d_K$ and let $\alpha_j\ge0$ with $\sum_{j=1}^N\alpha_j=1$. Define
$b=\sum_{j=1}^{N}\alpha_jv_j,
r=\sum_{j=1}^{N}\alpha_j\sqrt{1-K\|v_j\|^2}$ and the corresponding \PV{} midpoint aggregation is
\begin{equation}
\operatorname{Mid}^{\PV}_{K}\bigl(\{(v_j,\alpha_j)\}_{j=1}^{N}\bigr)
=\frac{b}{\sqrt{r^2+K\|b\|^2}}.
\label{eq:pv_midpoint_closed}
\end{equation}
\end{theorem}

\Cref{thm:pv_midpoint} is the \PV{} form of the Lorentzian midpoint: $r$ corresponds to the weighted time-like component and $b$ to the spatial component, so aggregation stays in \PV{} geometry rather than reverting to a Euclidean weighted sum.

\textbf{Attention weights and aggregation.}
The full attention variant applies a row-wise softmax to the Busemann logits and midpoint-aggregates values:
\begin{equation}
\alpha_{ij}
=\frac{\exp(S_{ij}^{\PV}/\sigma+b_{ij})}
{\sum_{\ell}\exp(S_{i\ell}^{\PV}/\sigma+b_{i\ell})},
\qquad
z_i=\operatorname{Mid}^{\PV}_{K}\bigl(\{(v_j,\alpha_{ij})\}_{j=1}^{N}\bigr),
\end{equation}
where $\sigma>0$ is the attention scale and $b_{ij}$ denotes an optional mask, edge bias, or positional bias. The medium-scale experiments use this full \PV{} Busemann attention path.

\subsection{Point-to-hyperplane transformation for keys}
\label{subsec:pv_pt2hyperplane_key}

The Busemann similarity in \Cref{subsec:pv_busemann_attention} requires queries and values to remain point-valued \PV{} representations, while keys must be represented as directions in the origin tangent space $T_0\mathcal{PV}^{m}_K$. Mapping every key by a logarithmic map at the origin would reintroduce a tangent-chart detour. We therefore construct a direct \PV{} point-to-hyperplane transformation: each output component is a signed distance-like margin to a learned \PV{} hyperplane, and stacking these margins yields the tangent key.

The Euclidean fully connected layer is defined as $y=Wx+b$ with $W\in\mathbb{R}^{m\times n}$ and $b\in\mathbb{R}^{m}$. Shimizu et al.~\citep[Sec.~3.2]{HNN++} and Chen et al.~\citep[Sec.~3.1]{chen2025building} interpret Euclidean FC layers through signed distances~\cite{shi2026manifold}: we write each output component as $v_\ell(x)=\langle w_\ell,x\rangle-b_\ell$ and view it as the signed distance from $y$ to the axis-aligned hyperplane
\begin{equation}
H_{e^{(\ell)},0}=\{y\in\mathbb{R}^{m}:\langle e^{(\ell)},y-0\rangle=0\},
\qquad
\operatorname{sign}\!\bigl(\langle e^{(\ell)},y-0\rangle\bigr)d\bigl(y,H_{e^{(\ell)},0}\bigr)
=v_\ell(x),
\label{eq:pv_pt2h_def}
\end{equation}
where $e^{(\ell)}$ is the $\ell$th standard basis vector. The left-hand side of \Cref{eq:pv_pt2h_def} is exactly the $\ell$th output component.

Keeping the left-hand side and replacing the Euclidean affine score by a \PV{} signed hyperplane margin extends this formulation to a \PV{}-to-tangent key layer, analogous to Lorentz point-to-hyperplane layers based on hyperbolic MLR logits~\citep[Eq.~12]{bdeir2023fully}. The \PV{} counterpart uses the signed-margin construction of \PVNN{}~\citep{chen2026proper}, reviewed in \Cref{app:pv_key_transform}, and gives the following key map.

\begin{theorem}[PV point-to-hyperplane key map]
\label{thm:pv_point_to_hyperplane_key}
Let $K<0$ and let $x\in\mathcal{PV}^{n}_K$. For each output component $\ell=1,\ldots,m$, take unconstrained parameters $z_\ell\in\mathbb{R}^{n}\setminus\{0\}$ and $r_\ell\in\mathbb{R}$, and set
\begin{equation}
p_\ell=\operatorname{Exp}_{0}\left(r_\ell\frac{z_\ell}{\|z_\ell\|}\right),
\qquad
a_\ell=\operatorname{PT}_{0\to p_\ell}(z_\ell).
\end{equation}
Define $\Psi^{\PV}_{\theta}:\mathcal{PV}^{n}_K\to T_0\mathcal{PV}^{m}_K\simeq\mathbb{R}^{m}$ by
$\Psi^{\PV}_{\theta}(x)
=\bigl[\psi_1(x),\ldots,\psi_m(x)\bigr]^\top,$
where
\begin{equation}
\psi_\ell(x)
=\frac{\|z_\ell\|}{\sqrt{-K}}
\operatorname{arsinh}\left(
\cosh(\sqrt{-K} r_\ell)\frac{\sqrt{-K}}{\|z_\ell\|}\langle x,z_\ell\rangle
-\sinh(\sqrt{-K} r_\ell)\sqrt{1-K\|x\|^2}
\right).
\label{eq:pv_tangent_key_component}
\end{equation}
Then $\Psi^{\PV}_{\theta}(x)$ is the vector of signed \PV{} point-to-hyperplane margins associated with the learned hyperplanes $\mathcal{H}^{\PV}_{a_\ell,p_\ell}$, represented as an origin tangent-space key. Moreover, as $K\to0^-$,
\begin{equation}
\Psi^{\PV}_{\theta}(x)\to Zx+b,
\qquad
b_\ell=-r_\ell\|z_\ell\|,
\end{equation}
where the rows of $Z\in\mathbb{R}^{m\times n}$ are $z_\ell^\top$.
\end{theorem}

The proof is given in \Cref{app:proof_pv_point_to_hyperplane_key}. The output is not a point in $\mathcal{PV}^{m}_K$: it is a tangent vector in $T_0\mathcal{PV}^{m}_K$ whose components are signed hyperplane margins. This distinction is necessary because the Busemann score in \Cref{thm:pv_busemann_function} uses keys as directions. For a token or node representation $u_i$, the key used by \PVFormer{} is therefore $k_i=\Psi^{\PV}_{\theta}(u_i)\in T_0\mathcal{PV}^{m}_K$, $S_{ij}^{\PV}=s_{\PV}(q_i,k_j)$.

\subsection{PV linear attention}
\label{subsec:pv_linear_attention}

Full Busemann attention forms all pairwise query--key scores and therefore has quadratic cost in the number of tokens or nodes. This is not the appropriate path for large graphs such as ogbn-proteins, Amazon2M, and ogbn-arxiv. Following Hypformer's linear-attention logic, we avoid explicitly materializing the $N\times N$ attention matrix by changing the computation order from $(QK^\top)V$ to $Q(K^\top V)$. The difference is that \PVFormer{} keeps the value aggregation as a \PV{} midpoint rather than a Euclidean weighted sum.

In the linear branch, keys are no longer required to be Busemann directions in $T_0\mathcal{PV}^d_K$; instead, queries, keys, and values are point-valued \PV{} representations obtained by the homomorphism transformations in \Cref{subsec:pv_homomorphism}. For one head, let
\begin{equation}
q_i=\Phi_{R_Q}^{\PV}(u_i),
\qquad
k_i=\Phi_{R_K}^{\PV}(u_i),
\qquad
v_i=\Phi_{R_V}^{\PV}(u_i),
\qquad
R_Q,R_K,R_V\in SO(d_h).
\end{equation}

\textbf{Focused nonnegative feature map.}
As in Hypformer, we use a positive-focused feature map to sharpen linear attention. For an input row $e\in\mathbb{R}^{d_h}$, define
\begin{equation}
\widetilde{e}=\frac{\operatorname{ReLU}(e)+\varepsilon}{t},
\qquad
\varphi_p(e)
=\frac{\|\widetilde{e}\|_2}{\|\widetilde{e}^{\odot p}\|_2+\varepsilon}
\widetilde{e}^{\odot p},
\qquad p>1,
\end{equation}
where $t>0$ is a learnable scale. This gives nonnegative kernel weights
\begin{equation}
\lambda_{ij}=\varphi_p(q_i)^\top\varphi_p(k_j),
\qquad
\alpha_{ij}=\frac{\lambda_{ij}}{\sum_{\ell=1}^{N}\lambda_{i\ell}}.
\end{equation}
A direct midpoint implementation would insert these $\alpha_{ij}$ into \Cref{thm:pv_midpoint}, but doing so would still require all $N^2$ weights. The key observation is that the midpoint can be algebraically linearized.

\begin{theorem}[PV linear attention]
\label{thm:pv_linear_attention}
Let $K<0$, let $\lambda_{ij}=\varphi_p(q_i)^\top\varphi_p(k_j)\ge0$, and define $\rho_j=\sqrt{1-K\|v_j\|^2}$, $B_i=\sum_{j=1}^{N}\lambda_{ij}v_j$,$A_i=\sum_{j=1}^{N}\lambda_{ij}\rho_j$.
Then applying the \PV{} midpoint in \Cref{thm:pv_midpoint} with normalized weights $\alpha_{ij}=\lambda_{ij}/\sum_{\ell}\lambda_{i\ell}$ is equivalent to
\begin{equation}
o_i^{\mathrm{PVLin}}
=\frac{B_i}{\sqrt{A_i^2+K\|B_i\|^2}}.
\end{equation}
Moreover, as $K\to0^-$, the expression recovers Euclidean linear attention,
\begin{equation}
o_i^{\mathrm{PVLin}}
\longrightarrow
\frac{\sum_{j=1}^{N}\lambda_{ij}v_j}{\sum_{j=1}^{N}\lambda_{ij}}.
\end{equation}

\end{theorem}

The proof is given in \Cref{app:proof_linear_attention}.

\section{Experiments}

\subsection{Experimental setup}
In this section, we evaluate \PVFormer{} on graph benchmarks and semi-supervised image and text classification settings. Graphs provide a natural testbed because their connectivity often contains hierarchical and long-range structure, so we begin with small and medium graph benchmarks and then move to large-scale node property prediction. We further evaluate Mini-ImageNet and 20News-Groups, where input graphs are constructed by a $k$-nearest-neighbor procedure or removed entirely, to test whether the gains transfer beyond datasets with a native graph.

To ensure a controlled comparison, we follow the Hypformer protocol and table layout for preprocessing, data splits, metrics, and baselines \citep{hu2020open,chiang2019clustergcn,wu2023nodeformer,yang2024hypformer}. The reported \PVFormer{} rows are our results under the same protocol. Disease uses F1, ogbn-proteins uses ROC-AUC, and all other listed datasets use accuracy. The main architectural difference is the attention core: \PVFormer{} replaces Euclidean or Lorentz attention with \PV{} homomorphism transformations, \PV{} Busemann scoring, \PV{} midpoint aggregation, and the \PV{} linear attention variant used for large graphs.

\subsection{Main results}
\begin{table*}[t]
\centering
\caption{Testing results on large-scale node property prediction benchmarks and small and medium-sized graph benchmarks. F1-score is used for {\sc Disease}, ROC-AUC for ogbn-proteins, and Accuracy for all other datasets. OOM denotes out of memory during training or testing. The \PVFormer{} row is highlighted in gray, with best results shown in red bold.}
\label{tab:graph}
\setlength{\tabcolsep}{2.2pt}
\resizebox{\textwidth}{!}{%
\begin{tabular}{l|ccccc|ccc}
\toprule
& \multicolumn{5}{c|}{\textbf{Small/Medium}} & \multicolumn{3}{c}{\textbf{Large}} \\
\cmidrule(lr){2-6}\cmidrule(lr){7-9}
Method & {\sc Disease} & {\sc Airport} & {\sc Cora} & {\sc Citeseer} & {\sc PubMed} & ogbn-proteins & Amazon2M & ogbn-arxiv \\
\midrule
\#Nodes & $1{,}044$ & $2{,}665$ & $2{,}708$ & $3{,}327$ & $19{,}717$ & $132{,}534$ & $2{,}449{,}029$ & $169{,}343$ \\
\#Edges & $1{,}043$ & $2{,}664$ & $5{,}429$ & $4{,}732$ & $88{,}651$ & $39{,}561{,}252$ & $61{,}859{,}140$ & $1{,}166{,}243$ \\
\midrule\midrule
MLP & -- & -- & -- & -- & -- & $72.0\pm0.5$ & $63.5\pm0.1$ & $55.5\pm0.2$ \\
GCN~\citep{gcn2017} & $69.7\pm0.4$ & $81.4\pm0.6$ & $81.3\pm0.3$ & $71.6\pm0.4$ & $78.1\pm0.2$ & $72.5\pm0.4$ & $83.9\pm0.1$ & $71.7\pm0.3$ \\
GAT~\citep{velivckovic2017graph} & $70.4\pm0.4$ & $81.5\pm0.3$ & $83.0\pm0.7$ & $72.5\pm1.1$ & $79.0\pm0.3$ & -- & -- & -- \\
SGC~\citep{wu2019simplifying} & $69.1\pm0.6$ & $82.1\pm0.5$ & $80.1\pm0.2$ & $71.9\pm0.1$ & $78.7\pm0.1$ & $70.3\pm0.2$ & $81.2\pm0.1$ & $67.8\pm0.3$ \\
GCN-NSampler & -- & -- & -- & -- & -- & $73.5\pm1.3$ & $83.8\pm0.4$ & $68.5\pm0.2$ \\
GAT-NSampler & -- & -- & -- & -- & -- & $74.6\pm1.2$ & $85.2\pm0.3$ & $67.6\pm0.2$ \\
SIGN~\citep{frasca2020sign} & -- & -- & -- & -- & -- & $71.2\pm0.5$ & $81.0\pm0.3$ & $70.3\pm0.3$ \\
HGNN~\citep{liu2019hyperbolic} & $81.3\pm3.5$ & $84.7\pm1.0$ & $77.1\pm0.8$ & $70.0\pm1.0$ & $78.3\pm1.2$ & -- & -- & -- \\
HGCN~\citep{chami2019hyperbolic} & $88.2\pm0.7$ & $89.3\pm1.2$ & $76.5\pm0.6$ & $68.0\pm0.6$ & $78.0\pm1.0$ & -- & -- & -- \\
HGAT~\citep{chami2019hyperbolic} & $90.3\pm0.6$ & $89.6\pm1.0$ & $77.4\pm0.7$ & $68.6\pm0.3$ & $78.3\pm1.4$ & -- & -- & -- \\
GraphFormer~\citep{ying2021transformers} & $75.2\pm0.0$ & $88.1\pm1.2$ & $60.0\pm0.5$ & $61.4\pm0.6$ & $73.3\pm0.7$ & OOM & OOM & OOM \\
GraphTrans~\citep{wu2021representing} & $89.3\pm3.2$ & $94.3\pm0.6$ & $77.6\pm0.8$ & $65.1\pm1.4$ & $77.5\pm0.7$ & OOM & OOM & OOM \\
GraphGPS~\citep{rampek2022recipe} & $92.8\pm2.7$ & $94.5\pm0.9$ & $73.0\pm1.4$ & $62.0\pm1.5$ & $72.8\pm1.4$ & OOM & OOM & OOM \\
FPS-T~\citep{cho2023curve} & $88.6\pm0.9$ & $96.0\pm0.6$ & $82.3\pm0.7$ & $70.0\pm0.7$ & $78.5\pm0.6$ & -- & -- & -- \\
HAN~\citep{gulcehre2018hyperbolic} & $85.1\pm0.8$ & $92.9\pm0.6$ & $83.1\pm0.5$ & $72.4\pm0.5$ & $79.0\pm0.6$ & OOM & OOM & OOM \\
HNN++~\citep{HNN++} & $89.5\pm0.2$ & $92.3\pm0.3$ & $82.8\pm0.6$ & $71.5\pm1.3$ & $79.9\pm0.4$ & OOM & OOM & OOM \\
F-HNN~\citep{chen2021fully} & $92.3\pm1.1$ & $93.0\pm0.7$ & $81.0\pm0.7$ & $71.2\pm0.4$ & $77.5\pm0.8$ & OOM & OOM & OOM \\
NodeFormer~\citep{wu2023nodeformer} & $75.9\pm0.9$ & $80.2\pm0.6$ & $82.2\pm0.9$ & $72.5\pm1.1$ & $79.9\pm1.0$ & $77.5\pm1.2$ & $87.9\pm0.2$ & $59.9\pm0.4$ \\
SGFormer~\citep{wu2023simplifying} & $89.0\pm3.9$ & $92.9\pm0.5$ & $83.2\pm0.9$ & $72.2\pm0.3$ & $80.0\pm0.8$ & $79.5\pm0.3$ & $89.1\pm0.1$ & $72.4\pm0.3$ \\
\Hypformer{} & $93.0\pm0.7$ & $95.0\pm0.5$ & $85.0\pm0.3$ & $73.3\pm0.4$ & $81.3\pm0.3$ & $80.4\pm0.5$ & $89.4\pm0.3$ & $73.2\pm0.2$ \\
\midrule
\rowcolor{gray!20}\darkred{\textbf{\PVFormer{}}} & \darkred{$\mathbf{94.07}\pm\mathbf{0.44}$} & \darkred{$\mathbf{96.37}\pm\mathbf{0.27}$} & \darkred{$\mathbf{86.20}\pm\mathbf{0.18}$} & \darkred{$\mathbf{73.70}\pm\mathbf{0.20}$} & \darkred{$\mathbf{83.00}\pm\mathbf{0.24}$} & \darkred{$\mathbf{81.51}\pm\mathbf{0.30}$} & \darkred{$\mathbf{90.03}\pm\mathbf{0.19}$} & \darkred{$\mathbf{73.86}\pm\mathbf{0.21}$} \\ \bottomrule
\end{tabular}%
}
\vspace{-3mm}
\end{table*}

\textbf{Graph results.}
\Cref{tab:graph} combines the small/medium graph benchmarks and large-scale node property prediction benchmarks in one grouped table. On small and medium graphs, \PVFormer{} improves over the \Hypformer{} reference means on all five datasets, with the largest gains on Airport, Cora, and PubMed, and smaller but positive margins on Disease and Citeseer. On the large benchmarks, \PVFormer{} improves over \Hypformer{} on ogbn-proteins, Amazon2M, and ogbn-arxiv while retaining the scalable \PV{} linear attention path. These results indicate that the \PV{} attention pipeline improves accuracy across graph scales while preserving the efficiency required by large graphs.

\begin{table*}[t]
\centering
\caption{Experimental results on semi-supervised classification on Mini-ImageNet and 20News-Groups, where we use k-NN graphs with different values of $k$. The \PVFormer{} rows are highlighted in gray, with our results shown in red bold.}
\label{table:image_text}
\setlength{\tabcolsep}{3.2pt}
\resizebox{\textwidth}{!}{
\begin{tabular}{@{}lcccccccc@{}}
\toprule
\multirow{2}{*}{Method} & \multicolumn{4}{c}{Mini-ImageNet} & \multicolumn{4}{c}{20News-Groups} \\ \cmidrule(lr){2-5}\cmidrule(l){6-9}
                        & $k=5$ & $k=10$ & $k=15$ & $k=20$ & $k=5$ & $k=10$ & $k=15$ & $k=20$ \\ \midrule
GCN~\citep{gcn2017} & $84.86 \pm 0.42$ & $85.61 \pm 0.40$ & $85.93 \pm 0.59$ & $85.96 \pm 0.66$ & $65.98 \pm 0.68$ & $64.13 \pm 0.88$ & $62.95 \pm 0.70$ & $62.59 \pm 0.62$ \\
GAT~\citep{velivckovic2017graph} & $84.70 \pm 0.48$ & $85.24 \pm 0.42$ & $85.41 \pm 0.43$ & $85.37 \pm 0.51$ & $64.06 \pm 0.44$ & $62.51 \pm 0.71$ & $61.38 \pm 0.88$ & $60.80 \pm 0.59$ \\
DropEdge~\citep{rong2019dropedge} & $83.91 \pm 0.24$ & $85.35 \pm 0.44$ & $85.25 \pm 0.63$ & $85.81 \pm 0.65$ & $64.46 \pm 0.43$ & $64.01 \pm 0.42$ & $62.46 \pm 0.51$ & $62.68 \pm 0.71$ \\
IDGL~\citep{chen2020iterative} & $83.63 \pm 0.32$ & $84.41 \pm 0.35$ & $85.50 \pm 0.24$ & $85.66 \pm 0.42$ & $65.09 \pm 1.23$ & $63.41 \pm 1.26$ & $61.57 \pm 0.52$ & $62.21 \pm 0.79$ \\
LDS~\citep{franceschi2019learning} & OOM & OOM & OOM & OOM & $66.15 \pm 0.36$ & $64.70 \pm 1.07$ & $63.51 \pm 0.64$ & $63.51 \pm 1.75$ \\
NodeFormer~\citep{wu2023nodeformer} & $86.77 \pm 0.45$ & $86.74 \pm 0.23$ & $86.87 \pm 0.41$ & ${86.64 \pm 0.42}$ & $66.01 \pm 1.18$ & ${65.21 \pm 1.14}$ & ${64.69 \pm 1.31}$ & ${64.55 \pm 0.97}$ \\
SGFormer~\citep{wu2023simplifying} & $86.21 \pm 0.66$ & $86.46 \pm 0.61$ & $86.73 \pm 0.84$ & $86.76 \pm 0.72$ & $68.55 \pm 0.54$ & $67.96 \pm 0.68$ & $66.44 \pm 0.87$ & $65.46 \pm 0.59$ \\
NodeFormer w/o graph~\citep{wu2023simplifying} & \multicolumn{4}{c}{$87.46 \pm 0.36$} & \multicolumn{4}{c}{$64.71 \pm 1.33$} \\
SGFormer w/o graph~\citep{wu2023simplifying} & \multicolumn{4}{c}{$87.25 \pm 0.38$} & \multicolumn{4}{c}{$67.53 \pm 0.43$} \\
Hypformer~\citep{yang2024hypformer} & $87.36 \pm 0.73$ & $87.30 \pm 0.65$ & $87.41 \pm 0.59$ & $87.48 \pm 0.61$ & $68.21 \pm 0.78$ & $68.01 \pm 0.34$ & $66.87 \pm 0.30$ & $66.74 \pm 0.19$ \\
Hypformer w/o graph~\citep{yang2024hypformer} & \multicolumn{4}{c}{$87.73 \pm 0.63$} & \multicolumn{4}{c}{$67.73 \pm 0.23$} \\ \midrule
\rowcolor{gray!20} \darkred{\textbf{\PVFormer{}}} & \darkred{$\mathbf{88.31}\pm\mathbf{0.21}$} & \darkred{$\mathbf{88.38}\pm\mathbf{0.32}$} & \darkred{$\mathbf{88.51}\pm\mathbf{0.29}$} & \darkred{$\mathbf{88.76}\pm\mathbf{0.30}$} & \darkred{$\mathbf{71.99}\pm\mathbf{0.51}$} & \darkred{$\mathbf{70.62}\pm\mathbf{0.30}$} & \darkred{$\mathbf{70.33}\pm\mathbf{0.24}$} & \darkred{$\mathbf{70.16}\pm\mathbf{0.20}$} \\
\rowcolor{gray!20} \darkred{\textbf{\PVFormer{} w/o graph}} & \multicolumn{4}{>{\columncolor{gray!20}}c}{\darkred{$\mathbf{88.47}\pm\mathbf{0.30}$}} & \multicolumn{4}{>{\columncolor{gray!20}}c}{\darkred{$\mathbf{70.29}\pm\mathbf{0.15}$}} \\ \bottomrule
\end{tabular}%
}
\vspace{-3mm}
\end{table*}

\textbf{Text and vision results.}
\Cref{table:image_text} evaluates semi-supervised classification on Mini-ImageNet and 20News-Groups. Following graph Transformer protocols, we construct $k$-nearest-neighbor graphs for $k\in\{5,10,15,20\}$ and also report graph-free variants. This setup separates gains from the learned \PV{} attention block from gains that depend on a particular neighborhood construction. On 20News-Groups, \PVFormer{} substantially improves over the Hypformer reference across every $k$ and in the graph-free setting. The best gain appears at $k=5$, where accuracy rises from 68.21 to 71.99, and all four graph settings exceed the baseline by 1.5 points. The graph-free 20News-Groups result of 70.29 shows that the improvement is not tied to a single constructed graph. On Mini-ImageNet, \PVFormer{} is consistently above the Hypformer paper baseline, with the best graph result of 88.76 at $k=20$ and 88.47 without an explicit graph.

\subsection{Ablations}

\textbf{Curvature sensitivity.}
\Cref{tab:curvature_sensitivity} evaluates whether the \PVFormer{} results depend on a finely tuned curvature value. We keep the input \PV{} scale fixed and sweep the hidden/output curvature over different $K$ on Airport, Cora, Citeseer, and PubMed. We also include the Euclidean counterpart ($K=0$) under the same architecture. All tested negative-curvature settings outperform this baseline on all four datasets. Among the negative-curvature settings, the Highest Test span is at most 0.44 points, and it is below 0.25 points on Cora, Citeseer, and PubMed. This indicates that \PVFormer{} is not sensitive to the exact negative curvature within this range; choosing a reasonable \PV{} curvature is sufficient once the surrounding optimization hyperparameters are fixed.

\begin{table}[!htbp]
\centering
\caption{Curvature sensitivity of \PVFormer{} on small and medium graph benchmarks, including the Euclidean baseline ($K=0$). Span denotes the difference between the best and worst mean Highest Test scores for $K<0$.}
\label{tab:curvature_sensitivity}
\setlength{\tabcolsep}{4.2pt}
\resizebox{\linewidth}{!}{%
\begin{tabular}{lcccccc>{\columncolor{gray!20}}c}
\toprule
Dataset & \shortstack{$K=0$\\(Euclidean)} & $K=-0.5$ & $K=-1.0$ & $K=-2.0$ & $K=-3.0$ & $K=-4.0$ & Span \\ \midrule
{\sc Airport} & $95.02\pm0.11$ & $96.06\pm0.11$ & $96.25\pm0.11$ & $96.50\pm0.22$ & $96.06\pm0.29$ & $96.31\pm0.29$ & $0.44$ \\
{\sc Cora} & $85.27\pm0.15$ & $86.10\pm0.10$ & $86.07\pm0.15$ & $86.07\pm0.15$ & $86.10\pm0.10$ & $86.13\pm0.06$ & $0.06$ \\
{\sc Citeseer} & $73.13\pm0.06$ & $73.93\pm0.15$ & $73.70\pm0.20$ & $73.70\pm0.26$ & $73.77\pm0.15$ & $73.77\pm0.06$ & $0.23$ \\
{\sc PubMed} & $82.13\pm0.32$ & $83.27\pm0.38$ & $83.13\pm0.32$ & $83.13\pm0.23$ & $83.13\pm0.23$ & $83.17\pm0.21$ & $0.14$ \\ \bottomrule
\end{tabular}%
}
\end{table}

\textbf{Query/value map ablation.}
To isolate the role of the \PV{} homomorphic query and value maps, we replace only the $Q$ and $V$ transformations with the \PVNN{} fully connected layer~\citep{chen2026proper} while other settings are kept fixed. \Cref{tab:pvnn_fc_qv_ablation} reports the Highest Test under this controlled replacement. According to the table, PVNN-FC is lower than the homomorphic map on all four datasets, with drops of $0.87$, $0.33$, $0.30$, and $0.50$ points on Airport, Cora, Citeseer, and PubMed, respectively. Additional gradient diagnostics for these transformations and similarity functions are provided in \Cref{app:pvnn_gradient_ablation}.

\begin{table}[!htbp]
\centering
\caption{Ablation replacing \PVFormer{} homomorphic query/value maps with the \PVNN{} FC.}
\label{tab:pvnn_fc_qv_ablation}
\setlength{\tabcolsep}{4.0pt}
\resizebox{0.75\linewidth}{!}{%
\begin{tabular}{lcccc}
\toprule
Q/V map & {\sc Airport} & {\sc Cora} & {\sc Citeseer} & {\sc PubMed} \\ \midrule
\rowcolor{gray!20}Homomorphic & $96.37\pm0.27$ & $86.20\pm0.18$ & $73.70\pm0.20$ & $83.00\pm0.24$ \\
PVNN-FC & $95.50\pm0.43$ & $85.87\pm0.25$ & $73.40\pm0.46$ & $82.50\pm0.30$ \\ \bottomrule
\end{tabular}%
}
\end{table}

\textbf{Similarity and key-map ablation.}
We next isolate the two geometry components inside full \PV{} attention: the Busemann similarity and the point-to-hyperplane key transformation. The distance-similarity variant in row 2 replaces the Busemann score with the negative \PV{} distance between the query point and a homomorphic key point. The logmap-key variant in row 3 keeps the Busemann score but replaces the point-to-hyperplane key map with the origin log map $\log_{0}(x)$. \Cref{tab:similarity_key_ablation} shows that the \PVFormer{} design is best on all four datasets. Replacing Busemann attention with distance-based similarity drops Airport, Cora, Citeseer, and PubMed by $6.10$, $2.03$, $0.93$, and $1.93$ points, respectively. Replacing the point-to-hyperplane key transformation with an origin log map yields a similar degradation: $6.48$, $2.03$, $0.77$, and $1.97$ points on the same datasets. The ablation rows also have much larger standard deviations on Cora, Citeseer, and PubMed, indicating less stable training.

\begin{table}[!htbp]
\centering
\caption{Similarity and key-map ablations on small and medium graph benchmarks.}
\label{tab:similarity_key_ablation}
\setlength{\tabcolsep}{4.0pt}
\resizebox{0.75\linewidth}{!}{%
\begin{tabular}{lcccc}
\toprule
Attention variant & {\sc Airport} & {\sc Cora} & {\sc Citeseer} & {\sc PubMed} \\ \midrule
\rowcolor{gray!20}\PVFormer{} & $96.37\pm0.27$ & $86.20\pm0.18$ & $73.70\pm0.20$ & $83.00\pm0.24$ \\
Distance similarity & $90.27\pm0.19$ & $84.17\pm1.33$ & $72.77\pm1.26$ & $81.07\pm2.08$ \\
Logmap key & $89.89\pm0.20$ & $84.17\pm1.33$ & $72.93\pm1.35$ & $81.03\pm2.04$ \\ \bottomrule
\end{tabular}%
}
\end{table}

\begin{figure}[!t]
\centering
\includegraphics[width=\linewidth]{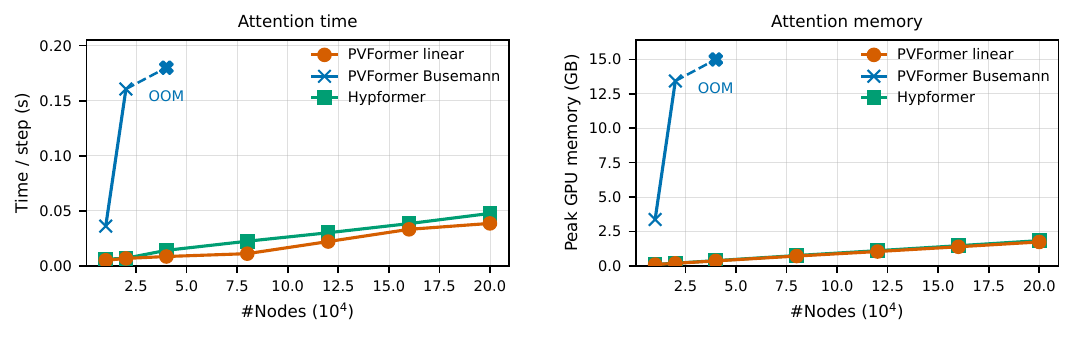}
\caption{Scalability of \PV{} linear attention and full \PV{} Busemann attention on an A100, with \Hypformer{} linear attention included as a reference. We report forward/backward time per attention step and peak GPU memory as the number of input nodes increases. Full Busemann attention is quadratic in $N$ and becomes infeasible at larger node counts, while both linear-attention branches remain feasible up to 200K nodes.}
\label{fig:pv_attention_scalability}
\end{figure}
\subsection{Attention Efficiency and Scalability}
We compare the two \PVFormer{} attention mechanisms while varying the number of input nodes, and include \Hypformer{} linear attention as a reference line. We fix the hidden dimension to 128, use one attention head, and measure forward/backward time per attention step and peak GPU memory. The controlled \PVFormer{} comparison changes only the attention mechanism: as shown in \Cref{fig:pv_attention_scalability}, \PV{} linear attention avoids materializing pairwise scores and has complexity $O(Nd_h^2)$, whereas full \PV{} Busemann attention forms an $N\times N$ score matrix and has complexity $O(N^2d_h)$. Furthermore, compared with \Hypformer{} linear attention, \PV{} linear attention consistently requires less time per step, indicating higher efficiency under the same input setting. On real large-scale graphs, the attention-branch time per step decreases by $11.7\%$ to $33.9\%$ in the tested settings (\Cref{app:real_graph_efficiency}).

\section{Conclusion}
We introduced \PVFormer{}, to the best of our knowledge, the first Transformer architecture operating natively in proper velocity space. By aligning the transformation, similarity, and aggregation primitives of self-attention with \PV{} geometry, \PVFormer{} combines \PV{} homomorphism transformations, \PV{} Busemann attention, a \PV{} point-to-hyperplane key map, \PV{} midpoint aggregation, and scalable \PV{} linear attention. Experiments under Hypformer protocols show consistent improvements on small and medium graph benchmarks, stable curvature sensitivity, controlled query/value, similarity, and key-map ablations, strong gains on the large-scale graph benchmarks, and substantial improvements on 20News-Groups, while Mini-ImageNet remains above baseline as well. \PVFormer{} may benefit hierarchical modeling, but requires careful validation before high-stakes use. As future work, we plan to extend our \PVFormer{} to large language models, exploiting PV space for large-scale deep learning.

\section*{Acknowledgements}
This work was supported by the EU Horizon Europe projects ELIAS (No. 101120237) and ELLIOT (No. 101214398), and by the FIS project GUIDANCE (No. FIS2023-03251). The authors gratefully acknowledge CINECA and the EuroHPC Joint Undertaking for providing access to the Leonardo supercomputer at CINECA, Italy. The authors declare no competing interests.
\bibliography{refs}

\input{appendix}

\section*{NeurIPS Paper Checklist}

\begin{enumerate}

\item {\bf Claims}
    \item[] Question: Do the main claims made in the abstract and introduction accurately reflect the paper's contributions and scope?
    \item[] Answer: \answerYes{}.
    \item[] Justification: The abstract and introduction state the native PV-space Transformer claim with the qualifier ``to the best of our knowledge'' and distinguish positive table results from the remaining Mini-ImageNet and reproducibility caveats.
\item {\bf Limitations}
    \item[] Question: Does the paper discuss the limitations of the work performed by the authors?
    \item[] Answer: \answerYes{}.
    \item[] Justification: See Appendix~\ref{app:limitations}.

\item {\bf Theory assumptions and proofs}
    \item[] Question: For each theoretical result, does the paper provide the full set of assumptions and a complete (and correct) proof?
    \item[] Answer: \answerYes{}.
    \item[] Justification: In Appendix D.

\item {\bf Experimental result reproducibility}
    \item[] Question: Does the paper fully disclose all the information needed to reproduce the main experimental results of the paper to the extent that it affects the main claims and/or conclusions of the paper (regardless of whether the code and data are provided or not)?
    \item[] Answer: \answerYes{}.
    \item[] Justification: The code repository is \url{https://github.com/Longchentong/PVFormer}.

\item {\bf Open access to data and code}
    \item[] Question: Does the paper provide open access to the data and code, with sufficient instructions to faithfully reproduce the main experimental results, as described in supplemental material?
    \item[] Answer: \answerYes{}.
    \item[] Justification: The code repository is \url{https://github.com/Longchentong/PVFormer}.

\item {\bf Experimental setting/details}
    \item[] Question: Does the paper specify all the training and test details (e.g., data splits, hyperparameters, how they were chosen, type of optimizer) necessary to understand the results?
    \item[] Answer: \answerYes{}.
    \item[] Justification: In Appendix E.

\item {\bf Experiment statistical significance}
    \item[] Question: Does the paper report error bars suitably and correctly defined or other appropriate information about the statistical significance of the experiments?
    \item[] Answer: \answerYes{}.
    \item[] Justification: The main \PVFormer{} table rows report mean and standard deviation.

\item {\bf Experiments compute resources}
    \item[] Question: For each experiment, does the paper provide sufficient information on the computer resources (type of compute workers, memory, time of execution) needed to reproduce the experiments?
    \item[] Answer: \answerYes{}.
    \item[] Justification: A100 GPU.

\item {\bf Code of ethics}
    \item[] Question: Does the research conducted in the paper conform, in every respect, with the NeurIPS Code of Ethics \url{https://neurips.cc/public/EthicsGuidelines}?
    \item[] Answer: \answerYes{}.
    \item[] Justification: The work uses public benchmark-style datasets and standard representation-learning experiments, and the manuscript avoids deanonymizing information.

\item {\bf Broader impacts}
    \item[] Question: Does the paper discuss both potential positive societal impacts and negative societal impacts of the work performed?
    \item[] Answer: \answerYes{}.
    \item[] Justification: In conclusion.
\item {\bf Safeguards}
    \item[] Question: Does the paper describe safeguards that have been put in place for responsible release of data or models that have a high risk for misuse (e.g., pre-trained language models, image generators, or scraped datasets)?
    \item[] Answer: \answerNA{}.
    \item[] Justification: The work does not release high-risk generative models, scraped datasets, or dual-use systems.
\item {\bf Licenses for existing assets}
    \item[] Question: Are the creators or original owners of assets (e.g., code, data, models), used in the paper, properly credited and are the license and terms of use explicitly mentioned and properly respected?
    \item[] Answer: \answerYes{}.
    \item[] Justification: The paper cites benchmark and method papers where available, but the paper adopts CC BY 4.0.
\item {\bf New assets}
    \item[] Question: Are new assets introduced in the paper well documented and is the documentation provided alongside the assets?
    \item[] Answer: \answerYes{}.
    \item[] Justification: The paper introduces a method and experimental results but does not release a new dataset or standalone asset package.
\item {\bf Crowdsourcing and research with human subjects}
    \item[] Question: For crowdsourcing experiments and research with human subjects, does the paper include the full text of instructions given to participants and screenshots, if applicable, as well as details about compensation (if any)?
    \item[] Answer: \answerNA{}.
    \item[] Justification: The work does not involve crowdsourcing or human-subject experiments.

\item {\bf Institutional review board (IRB) approvals or equivalent for research with human subjects}
    \item[] Question: Does the paper describe potential risks incurred by study participants, whether such risks were disclosed to the subjects, and whether Institutional Review Board (IRB) approvals (or an equivalent approval/review based on the requirements of your country or institution) were obtained?
    \item[] Answer: \answerNA{}.
    \item[] Justification: The work does not involve human-subject experiments.
\item {\bf Declaration of LLM usage}
    \item[] Question: Does the paper describe the usage of LLMs if it is an important, original, or non-standard component of the core methods in this research? Note that if the LLM is used only for writing, editing, or formatting purposes and does \emph{not} impact the core methodology, scientific rigor, or originality of the research, declaration is not required.
    \item[] Answer: \answerNA{}.
    \item[] Justification: LLM usage is not part of the core \PVFormer{} method or experimental claims in this work.

\end{enumerate}

\end{document}

%% file: appendix.tex
\clearpage

\appendix
{\hypersetup{linkcolor=black}
\providecommand{\authcount}[1]{}
\providecommand{\startcontents}[1][]{}
\providecommand{\printcontents}[4][]{}
\startcontents[sections]
\section*{Appendix Contents}
\printcontents[sections]{l}{1}{\setcounter{tocdepth}{2}}
}
\section{Notation}

\Cref{tab:app_notation} summarizes the notation used throughout the main paper and appendix. We use lowercase letters for vectors in the proper velocity and Poincar\'e models, and uppercase letters for ambient Lorentz vectors when the context is clear.

\begin{table}[ht]
\centering
\caption{Summary of notation used throughout the main paper and appendix.}
\label{tab:app_notation}
\small
\renewcommand{\arraystretch}{1.08}
\begin{tabular}{@{}p{0.28\linewidth}p{0.64\linewidth}@{}}
\toprule
Notation & Description \\
\midrule
$K<0$ & Constant sectional curvature; all formulas keep $K$ negative. \\
$\mathbb{R}^d$ & $d$-dimensional Euclidean space. \\
$\langle\cdot,\cdot\rangle$ and $\|\cdot\|$ & Euclidean inner product and norm. \\
$(\mathcal{M},g)$ & Riemannian manifold with metric $g$. \\
$T_x\mathcal{M}$, $g_x(\cdot,\cdot)$, and $\|\cdot\|_x$ & Tangent space at $x$, Riemannian metric, and induced tangent norm. \\
$d_{\mathcal{M}}$, $\operatorname{Exp}_x$, $\operatorname{Log}_x$, and $\operatorname{PT}_{x\to y}$ & Geodesic distance, exponential map, logarithmic map, and parallel transport. \\
$\mathcal{PV}_K^d=\mathbb{R}^d$ & $d$-dimensional proper velocity model equipped with the \PV{} metric. \\
$\beta_x$ and $\rho_x$ & Relativistic factors $\beta_x=(1-K\|x\|^2)^{-1/2}$ and $\rho_x=\beta_x^{-1}$. \\
$\oplus_U$, $\ominus_Ux$, and $t\otimes_Ux$ & \PV{} gyroaddition, inverse, and scalar gyromultiplication~\citep{ungar2022analytic,chen2026proper}. \\
$\mathbb{L}_K^d$ and $\mathbb{P}_K^d$ & Lorentz model and Poincar\'e ball model with curvature $K<0$. \\
$\langle X,Y\rangle_{\mathcal{L}}$ and $\|V\|_{\mathcal{L}}$ & Lorentz inner product and the induced norm on tangent vectors. \\
$X=(X_0,X_s)$ and $\bar{0}$ & Time-like/spatial decomposition of a Lorentz vector and the Lorentz origin $((-K)^{-1/2},0,\ldots,0)$. \\
$N$, $h$, and $d_h$ & Number of tokens or nodes, attention-head index, and per-head hidden dimension. \\
$u_i$, $q_i$, $k_i$, and $v_i$ & \PV{} hidden state, point-valued query, direction-like key, and point-valued value in one attention head. \\
$\hat{k}_i$, $S_{ij}^{\PV}$, and $\alpha_{ij}$ & Normalized key direction, \PV{} attention logit, and normalized attention weight. \\
$\operatorname{Mid}_K^{\PV}$ & \PV{} midpoint aggregation operator. \\
$\gamma$ and $B^\gamma$ & Geodesic ray and its Busemann function. \\
$B_K^{\PV,u}$ & \PV{} Busemann function associated with unit tangent direction $u\in T_0\mathcal{PV}_K^d$. \\
$\mathcal{H}_{a,p}^{\PV}$ & \PV{} hyperplane with base point $p$ and normal $a\in T_p\mathcal{PV}_K^d$. \\
$\Psi_\theta^{\PV}$ & Learned point-to-hyperplane map from a \PV{} point to an origin-tangent key vector. \\
\bottomrule
\end{tabular}
\end{table}

\section{Preliminaries}

\subsection{Riemannian notation and operations}

For completeness, we spell out the Riemannian conventions compressed in the main text. A Riemannian manifold $(\mathcal{M},g)$ is a smooth manifold equipped with a smoothly varying inner product $g_x:T_x\mathcal{M}\times T_x\mathcal{M}\to\mathbb{R}$ on each tangent space. The induced norm is $\|v\|_x=\sqrt{g_x(v,v)}$. Geodesics generalize straight lines, and the Riemannian distance $d_{\mathcal{M}}(x,y)$ is the length of the shortest geodesic between $x$ and $y$ when it exists. The exponential map $\operatorname{Exp}_x$ moves from $x$ along a geodesic with a specified initial tangent vector, while $\operatorname{Log}_x$ is its local inverse. Parallel transport $\operatorname{PT}_{x\to y}$ moves tangent vectors along a geodesic while preserving the Riemannian metric. A smooth bijection between two Riemannian manifolds is an isometry if its differential preserves tangent inner products.

\subsection{Hyperbolic models and isometries}

Hyperbolic space of curvature $K<0$ can be represented by several equivalent models. The Lorentz model summarized in \Cref{tab:app_notation} realizes it as the upper sheet of the two-sheeted hyperboloid. Its geodesic distance is
\begin{equation}
d_{\mathbb{L}}(X,Y)=\frac{1}{\sqrt{-K}}\operatorname{arcosh}\!\left(K\langle X,Y\rangle_{\mathcal{L}}\right),
\qquad X,Y\in\mathbb{L}_K^d,
\end{equation}
which is the form used by Lorentz hyperbolic Transformers~\citep{yang2024hypformer} and Busemann-based hyperbolic learning~\citep{atigh2021hyperbolic}.

The \PVNN{} paper shows that the PV model is isometric to both the Poincar\'e and Lorentz models~\citep{chen2026proper}. The PV--Poincar\'e maps are
\begin{equation}
\pi_{\mathcal{PV}\to\mathbb{P}}(x)
=\frac{\beta_x}{1+\beta_x}x,
\qquad
\pi_{\mathbb{P}\to\mathcal{PV}}(y)=2\gamma_y^2y,
\qquad
\gamma_y=\frac{1}{\sqrt{1+K\|y\|^2}}.
\label{eq:app_pv_poincare_maps}
\end{equation}
The PV--Lorentz maps are especially simple:
\begin{equation}
\iota_{\mathcal{PV}\to\mathbb{L}}(x)
=\left(\sqrt{\|x\|^2-\frac{1}{K}},\;x\right),
\qquad
\pi_{\mathbb{L}\to\mathcal{PV}}(X_0,X_s)=X_s.
\label{eq:app_pv_lorentz_maps}
\end{equation}
Thus the PV representation stores exactly the spatial part of the Lorentz point, while the time-like component is recovered from the hyperboloid constraint. This observation is used throughout \PVFormer{}: operations derived in Lorentz space can be pulled back to the unconstrained PV model, avoiding a bounded Poincar\'e domain and avoiding explicit hyperboloid projection.

\subsection{Lorentz model operations}

For $X\in\mathbb{L}_K^d$, the tangent space and induced Riemannian metric are
\begin{equation}
T_X\mathbb{L}_K^d=\{V\in\mathbb{R}^{d+1}:\langle V,X\rangle_{\mathcal{L}}=0\},
\qquad
\langle U,V\rangle_X=\langle U,V\rangle_{\mathcal{L}}.
\end{equation}
For $Z\in T_X\mathbb{L}_K^d$ and $Y\in\mathbb{L}_K^d$, let
$\alpha=\sqrt{-K}\|Z\|_X$ and $\beta=K\langle X,Y\rangle_{\mathcal{L}}$. The exponential and logarithmic maps are
\begin{align}
\operatorname{Exp}^{\mathbb{L}}_X(Z)
&=\cosh(\alpha)X+\frac{\sinh(\alpha)}{\alpha}Z,
\\
\operatorname{Log}^{\mathbb{L}}_X(Y)
&=\frac{\operatorname{arcosh}(\beta)}{\sqrt{\beta^2-1}}(Y-\beta X),
\end{align}
with the usual continuous extension when $Z=0$ or $Y=X$. At the origin $\bar{0}=((-K)^{-1/2},0)$, identifying $T_{\bar{0}}\mathbb{L}_K^d$ with spatial vectors $z\in\mathbb{R}^d$ gives
\begin{align}
\operatorname{Exp}^{\mathbb{L}}_{\bar{0}}(z)
&=\frac{1}{\sqrt{-K}}\left(\cosh(\sqrt{-K}\|z\|),\;\sinh(\sqrt{-K}\|z\|)\frac{z}{\|z\|}\right),
\\
\operatorname{Log}^{\mathbb{L}}_{\bar{0}}(Y)
&=\left(0,\;\frac{\operatorname{arcosh}(\sqrt{-K}Y_0)}{\sqrt{(-K)Y_0^2-1}}Y_s\right),
\end{align}
again with the zero-vector convention. Parallel transport along the geodesic from $X$ to $Y$ is
\begin{equation}
\operatorname{PT}^{\mathbb{L}}_{X\to Y}(V)
=V+\frac{\langle Y,V\rangle_{\mathcal{L}}}{-1/K-\langle X,Y\rangle_{\mathcal{L}}}(X+Y),
\end{equation}
with the identity map when $X=Y$.

Lorentz transformations are matrices $A\in\mathbb{R}^{(d+1)\times(d+1)}$ satisfying $A^\top J A=J$ for $J=\operatorname{diag}(-1,1,\ldots,1)$. The time-orientation-preserving subgroup maps the upper sheet to itself, and the origin-preserving rotations have the form $\operatorname{diag}(1,R)$ with $R\in SO(d)$. This is the Lorentz transformation that becomes $x\mapsto Rx$ after projection to the PV model.

For a unit spatial direction $u\in\mathbb{S}^{d-1}$, the origin-centered geodesic ray is
\begin{equation}
\Gamma_u(t)=\left((-K)^{-1/2}\cosh(\sqrt{-K}t),\;(-K)^{-1/2}\sinh(\sqrt{-K}t)u\right).
\end{equation}
The corresponding Lorentz Busemann function is
\begin{equation}
B_{\mathbb{L},K}^{u}(X)
=\lim_{t\to\infty}\bigl(d_{\mathbb{L}}(X,\Gamma_u(t))-t\bigr)
=\frac{1}{\sqrt{-K}}\log\!\left(\sqrt{-K}X_0-\sqrt{-K}\langle X_s,u\rangle\right).
\label{eq:app_lorentz_busemann}
\end{equation}

\subsection{Proper velocity geometry}

The PV metric is
\begin{equation}
g_x(u,v)=\langle u,v\rangle+K\beta_x^2\langle x,u\rangle\langle x,v\rangle,
\qquad u,v\in T_x\mathcal{PV}_K^d\simeq\mathbb{R}^d.
\end{equation}
Because $K<0$, $1-K\|x\|^2>0$ for every $x\in\mathbb{R}^d$, so the model domain is all of $\mathbb{R}^d$. The associated PV gyrovector operations are
\begin{align}
x\oplus_U y
&=x+y+\left(\frac{1-\beta_y}{\beta_y}
-K\frac{\beta_x}{1+\beta_x}\langle x,y\rangle\right)x,
\\
t\otimes_U x
&=\frac{\sinh\!\left(t\operatorname{arsinh}(\sqrt{-K}\|x\|)\right)}{\sqrt{-K}\|x\|}x,
\qquad t\otimes_U0=0.
\label{eq:app_pv_scalar}
\end{align}
These operations reduce to ordinary addition and scalar multiplication as $K\to0^-$ and provide the algebraic structure used by PV homomorphism transformations.

PV Riemannian operations can be obtained by transferring the corresponding Poincar\'e operations through the isometry in \Cref{eq:app_pv_poincare_maps}. Let $\pi=\pi_{\mathcal{PV}\to\mathbb{P}}$ and let $d_x\pi$ be its differential. For $x,y\in\mathcal{PV}_K^d$ and $v\in T_x\mathcal{PV}_K^d$, define
\begin{equation}
z=(\ominus_U x)\oplus_U y,
\qquad \bar{z}=\pi(z).
\end{equation}
Then the distance is
\begin{equation}
d_{\PV}(x,y)=\frac{2}{\sqrt{-K}}\tanh^{-1}\!\left(\sqrt{-K}\left\|\pi\!\left((\ominus_U x)\oplus_U y\right)\right\|\right).
\end{equation}
The logarithmic map can be written as
\begin{equation}
\operatorname{Log}_x(y)=\sigma(x,y)z+\tau(x,y)\langle x,z\rangle x,
\end{equation}
where
\begin{equation}
\sigma(x,y)=\frac{2}{\sqrt{-K}}\frac{\tanh^{-1}(\sqrt{-K}\|\bar{z}\|)}{\|z\|},
\qquad
\tau(x,y)=\frac{2\beta_x}{1+\beta_x}\frac{\sqrt{-K}\tanh^{-1}(\sqrt{-K}\|\bar{z}\|)}{\|z\|}.
\end{equation}
At the origin, the operators simplify to
\begin{align}
\operatorname{Exp}_0(v)
&=\frac{1}{\sqrt{-K}}\sinh(\sqrt{-K}\|v\|)\frac{v}{\|v\|},
\\
\operatorname{Log}_0(y)
&=\frac{1}{\sqrt{-K}}\operatorname{arsinh}(\sqrt{-K}\|y\|)\frac{y}{\|y\|},
\\
d_{\PV}(0,y)
&=\frac{1}{\sqrt{-K}}\operatorname{arsinh}(\sqrt{-K}\|y\|),
\end{align}
with the usual continuous extension at the zero vector. These identity-centered expressions are the forms used to interpret tangent directions and key vectors in \PVFormer{}.

\subsection{Busemann functions}

Let $(X,d)$ be a proper geodesic metric space and let $\gamma:[0,\infty)\to X$ be a unit-speed geodesic ray. The Busemann function associated with $\gamma$ is
\begin{equation}
B^\gamma(x)=\lim_{t\to\infty}\left(d(x,\gamma(t))-t\right).
\end{equation}
In Euclidean space, if $\gamma(t)=tv$ for a unit direction $v$, then
\begin{equation}
B^v(x)=-\langle x,v\rangle.
\end{equation}
Thus the Busemann function is a curvature-aware analogue of a directional inner product, which motivates Busemann-based hyperbolic similarity~\citep{bridson2013metric,atigh2021hyperbolic}.

The Lorentz closed form in \Cref{eq:app_lorentz_busemann} is the curvature-$K$ directional Busemann potential. Pulling it back through the PV--Lorentz isometry in \Cref{eq:app_pv_lorentz_maps} gives the PV form. For $q\in\mathcal{PV}_K^d$ and $\|u\|=1$,
\begin{equation}
B_{\PV,K}^{u}(q)
=\frac{1}{\sqrt{-K}}\log\!\left(\sqrt{1-K\|q\|^2}-\sqrt{-K}\langle q,u\rangle\right).
\end{equation}
For a nonzero key vector $k\in T_0\mathcal{PV}_K^d$, \PVFormer{} uses the scaled similarity
\begin{equation}
s_{\PV}(q,k)=-\|k\|B_{\PV,K}^{k/\|k\|}(q).
\label{eq:app_pv_busemann_similarity}
\end{equation}
As $K\to0^-$, \Cref{eq:app_pv_busemann_similarity} converges to $\langle q,k\rangle$, so the score recovers the Euclidean attention logit in the flat limit.

\section{PV Point-to-Hyperplane Keys}
\label{app:pv_key_transform}

This section expands the point-to-hyperplane key transformation used in \Cref{subsec:pv_pt2hyperplane_key}. The construction follows the PV hyperplane and PV MLR derivations in \PVNN{}~\citep{chen2026proper}, but the role of the output is different: \PVFormer{} uses the resulting components as tangent-space key directions for Busemann attention rather than as class logits.

\subsection{PV hyperplanes and signed margins}

For $p\in\mathcal{PV}_K^d$ and $a\in T_p\mathcal{PV}_K^d$, define the PV hyperplane
\begin{equation}
\mathcal{H}_{a,p}^{\PV}
=\left\{x\in\mathcal{PV}_K^d:
    g_p\left(\operatorname{Log}_p(x),a\right)=0
\right\}.
\label{eq:app_pv_hyperplane_full}
\end{equation}
Let $\pi=\pi_{\mathcal{PV}\to\mathbb{P}}$ be the PV--Poincar\'e isometry in \Cref{eq:app_pv_poincare_maps}. Since isometries preserve distances and map totally geodesic hyperplanes to totally geodesic hyperplanes, the PV point-to-hyperplane distance can be computed by moving the hyperplane to the Poincar\'e ball. Equivalently, the signed margin has the closed form
\begin{equation}
\operatorname{margin}_{a,p}(x)
=\frac{1}{\sqrt{-K}}\operatorname{arsinh}\!\left(
\frac{\sqrt{-K}\left\langle (\ominus_U p)\oplus_U x,\,d_p\pi(a)\right\rangle}
{\|d_p\pi(a)\|}
\right),
\label{eq:app_signed_margin_general}
\end{equation}
where the sign indicates on which side of $\mathcal{H}_{a,p}^{\PV}$ the point lies. This form is the PV analogue of the signed-distance interpretation of Euclidean affine components.

\subsection{Efficient reparameterization for keys}

A direct parameterization by $(p,a)$ is redundant and expensive for a multi-output key map. Following \PVNN{}, we use a unidirectional parameterization. For the $\ell$th output component, take $z_\ell\in\mathbb{R}^d\setminus\{0\}$ and $r_\ell\in\mathbb{R}$, and set
\begin{equation}
p_\ell=\operatorname{Exp}_0\!\left(r_\ell\frac{z_\ell}{\|z_\ell\|}\right),
\qquad
 a_\ell=\operatorname{PT}_{0\to p_\ell}(z_\ell).
\label{eq:app_key_params}
\end{equation}
The resulting signed component can be written without explicitly constructing $p_\ell$ or $a_\ell$:
\begin{equation}
\psi_\ell(x)
=\frac{\|z_\ell\|}{\sqrt{-K}}
\operatorname{arsinh}\!\left(
\cosh(\sqrt{-K}r_\ell)\frac{\sqrt{-K}}{\|z_\ell\|}\langle x,z_\ell\rangle
-\sinh(\sqrt{-K}r_\ell)\sqrt{1-K\|x\|^2}
\right).
\label{eq:app_pv_key_component}
\end{equation}
Stacking $m$ components defines the key map
\begin{equation}
\Psi_\theta^{\PV}(x)
=\left[\psi_1(x),\ldots,\psi_m(x)\right]^\top
\in T_0\mathcal{PV}_K^m\simeq\mathbb{R}^m.
\label{eq:app_key_map}
\end{equation}
For \PVFormer{}, $\Psi_\theta^{\PV}(u_i)$ is not interpreted as a point in $\mathcal{PV}_K^m$. It is the tangent key vector that determines the Busemann direction in \Cref{eq:pv_busemann_score}. This separation keeps query and value representations point-valued while giving the similarity stage the directional object required by Busemann scoring.

\subsection{Flat limit}
\label{app:key_flat_limit}

The key map recovers an affine Euclidean projection as curvature vanishes. Using the Taylor expansions
\begin{align}
\cosh(\sqrt{-K}r_\ell)&=1+O(K),
&
\sinh(\sqrt{-K}r_\ell)&=\sqrt{-K}r_\ell+O((-K)^{3/2}),
\\
\sqrt{1-K\|x\|^2}&=1+O(K).&&
\end{align}
the argument of $\operatorname{arsinh}$ in \Cref{eq:app_pv_key_component} is
\begin{equation}
\sqrt{-K}\left(\frac{\langle x,z_\ell\rangle}{\|z_\ell\|}-r_\ell\right)+O((-K)^{3/2}).
\end{equation}
Since $\operatorname{arsinh}(t)=t+O(t^3)$ near zero,
\begin{equation}
\lim_{K\to0^-}\psi_\ell(x)=\langle x,z_\ell\rangle-r_\ell\|z_\ell\|.
\end{equation}
Thus $\Psi_\theta^{\PV}$ is a curvature-aware replacement for an affine key projection, while preserving the usual Euclidean key layer in the flat limit.

\section{Proofs and Derivations}

\subsection{Proof of the PV isotropy homomorphism}
\label{app:proof_isotropy}

We prove \Cref{thm:pv_isotropy_homomorphism}. Let $R\in SO(d)$ and $\Phi_R^{\PV}(x)=Rx$. Since $R^\top R=I$, $\|Rx\|=\|x\|$ and $\beta_{Rx}=\beta_x$. For tangent vectors $u,v\in T_x\mathcal{PV}_K^d$,
\begin{align}
g_{Rx}(Ru,Rv)
&=\langle Ru,Rv\rangle
+K\beta_{Rx}^2\langle Rx,Ru\rangle\langle Rx,Rv\rangle \\
&=\langle u,v\rangle
+K\beta_x^2\langle x,u\rangle\langle x,v\rangle
=g_x(u,v),
\end{align}
so $\Phi_R^{\PV}$ is a Riemannian isometry.

For gyroaddition, write
\begin{equation}
A(x,y)=\frac{1-\beta_y}{\beta_y}-K\frac{\beta_x}{1+\beta_x}\langle x,y\rangle,
\qquad x\oplus_U y=x+y+A(x,y)x.
\end{equation}
Orthogonality gives $\beta_{Rx}=\beta_x$, $\beta_{Ry}=\beta_y$, and $\langle Rx,Ry\rangle=\langle x,y\rangle$, hence $A(Rx,Ry)=A(x,y)$. Therefore
\begin{equation}
R(x\oplus_U y)=Rx+Ry+A(x,y)Rx=Rx\oplus_U Ry.
\end{equation}
The scalar law follows from the radial form of $t\otimes_U x$ in \Cref{eq:app_pv_scalar} and the identity $\|Rx\|=\|x\|$:
\begin{equation}
R(t\otimes_U x)=t\otimes_U(Rx).
\end{equation}
This proves the isometry, gyroaddition, and scalar-compatibility claims.

\subsection{Derivation of the PV Busemann score}
\label{app:proof_pv_busemann}

We derive \Cref{thm:pv_busemann_function}. Let $q\in\mathcal{PV}_K^d$ and $\|u\|=1$. Under the PV--Lorentz isometry,
\begin{equation}
q\mapsto X_q=\left(\sqrt{\|q\|^2-\frac{1}{K}},q\right)
=\left(\frac{\sqrt{1-K\|q\|^2}}{\sqrt{-K}},q\right).
\end{equation}
The unit-speed geodesic ray from the origin in direction $u$ lifts to
\begin{equation}
\Gamma_u(t)=\left((-K)^{-1/2}\cosh(\sqrt{-K}t),
(-K)^{-1/2}\sinh(\sqrt{-K}t)u\right).
\end{equation}
Using the Lorentz distance,
\begin{equation}
d_{\mathbb{L}}(X_q,\Gamma_u(t))
=\frac{1}{\sqrt{-K}}\operatorname{arcosh}\!\left(K\langle X_q,\Gamma_u(t)\rangle_{\mathcal{L}}\right).
\end{equation}
As $t\to\infty$, the argument of $\operatorname{arcosh}$ is asymptotic to
\begin{equation}
\frac{e^{\sqrt{-K}t}}{2}\left(\sqrt{1-K\|q\|^2}-\sqrt{-K}\langle q,u\rangle\right).
\end{equation}
Since $\operatorname{arcosh}(s)=\log(2s)+o(1)$ as $s\to\infty$, subtracting $t$ yields
\begin{equation}
B_{\PV,K}^u(q)
=\frac{1}{\sqrt{-K}}\log\!\left(\sqrt{1-K\|q\|^2}-\sqrt{-K}\langle q,u\rangle\right).
\end{equation}
The logarithm is positive: with $h=\sqrt{-K}$, Cauchy--Schwarz gives $h\langle q,u\rangle\le h\|q\|<\sqrt{1+h^2\|q\|^2}$. Multiplying by $-\|k\|$ with $u=k/\|k\|$ gives the similarity in \Cref{eq:pv_busemann_similarity_closed}. For the flat limit, let $a=\langle q,u\rangle$. Then
\begin{equation}
\log\left(\sqrt{1+h^2\|q\|^2}-ha\right)=-ha+O(h^2),
\end{equation}
so $-\|k\|h^{-1}\log(\sqrt{1+h^2\|q\|^2}-ha)=\|k\|a+O(h)\to\langle q,k\rangle$.

\subsection{Proof of the PV point-to-hyperplane key map}
\label{app:proof_pv_point_to_hyperplane_key}

We prove \Cref{thm:pv_point_to_hyperplane_key}. For $p\in\mathcal{PV}_K^n$ and $a\in T_p\mathcal{PV}_K^n$, \Cref{eq:app_pv_hyperplane_full} defines the corresponding \PV{} hyperplane and \Cref{eq:app_signed_margin_general} gives its signed margin. The unidirectional parameterization in \Cref{eq:app_key_params} sets $p_\ell=\operatorname{Exp}_0(r_\ell z_\ell/\|z_\ell\|)$ and $a_\ell=\operatorname{PT}_{0\to p_\ell}(z_\ell)$. Substituting these parameters into the signed-margin formula and simplifying the PV--Poincar\'e differential yields the component expression in \Cref{eq:app_pv_key_component}, which matches \Cref{eq:pv_tangent_key_component}. Stacking the $m$ components gives \Cref{eq:app_key_map}, hence $\Psi_\theta^{\PV}(x)$ is exactly the vector of signed \PV{} point-to-hyperplane margins represented in $T_0\mathcal{PV}_K^m\simeq\mathbb{R}^m$. The flat-limit statement follows from the Taylor expansion in \Cref{app:key_flat_limit}.

\subsection{Proof of PV midpoint aggregation}

We prove \Cref{thm:pv_midpoint}. Let $v_j\in\mathcal{PV}_K^d$ and lift each value to the Lorentz model:
\begin{equation}
V_j=\left(\sqrt{\|v_j\|^2-\frac{1}{K}},v_j\right)
=\left(\frac{\sqrt{1-K\|v_j\|^2}}{\sqrt{-K}},v_j\right).
\end{equation}
The Lorentzian midpoint first forms $\widetilde{V}=\sum_j\alpha_jV_j$ and then normalizes it back to the hyperboloid. Its spatial part before normalization is $b=\sum_j\alpha_jv_j$, and its time component is $(-K)^{-1/2}r$ with
\begin{equation}
r=\sum_j\alpha_j\sqrt{1-K\|v_j\|^2}.
\end{equation}
The Lorentz norm of $\widetilde{V}$ is
\begin{equation}
\langle\widetilde{V},\widetilde{V}\rangle_{\mathcal{L}}
=-\frac{r^2}{-K}+\|b\|^2
=\frac{r^2+K\|b\|^2}{K}.
\end{equation}
The normalization factor that restores curvature $K$ is therefore $(r^2+K\|b\|^2)^{-1/2}$ on the spatial part, which gives
\begin{equation}
\operatorname{Mid}^{\PV}_{K}\bigl(\{(v_j,\alpha_j)\}_{j=1}^N\bigr)
=\frac{b}{\sqrt{r^2+K\|b\|^2}}.
\end{equation}

\subsection{Proof of PV linear attention}
\label{app:proof_linear_attention}

For \Cref{thm:pv_linear_attention}, let $\lambda_{ij}\ge0$ be the unnormalized kernel weights and $Z_i=\sum_j\lambda_{ij}$. The normalized midpoint weights are $\alpha_{ij}=\lambda_{ij}/Z_i$. Substituting these weights into \Cref{thm:pv_midpoint} gives
\begin{equation}
b_i=\sum_j\alpha_{ij}v_j=\frac{1}{Z_i}\sum_j\lambda_{ij}v_j=Z_i^{-1}B_i,
\qquad
r_i=\sum_j\alpha_{ij}\rho_j=\frac{1}{Z_i}\sum_j\lambda_{ij}\rho_j=Z_i^{-1}A_i,
\end{equation}
where $B_i=\sum_j\lambda_{ij}v_j$, $A_i=\sum_j\lambda_{ij}\rho_j$, and $\rho_j=\sqrt{1-K\|v_j\|^2}$. Inserting these expressions into the closed-form midpoint in \Cref{eq:pv_midpoint_closed} yields
\begin{align}
o_i^{\mathrm{PVLin}}
&=\frac{b_i}{\sqrt{r_i^2+K\|b_i\|^2}} \\
&=\frac{Z_i^{-1}B_i}{\sqrt{Z_i^{-2}A_i^2+KZ_i^{-2}\|B_i\|^2}} \\
&=\frac{Z_i^{-1}B_i}{Z_i^{-1}\sqrt{A_i^2+K\|B_i\|^2}} \\
&=\frac{B_i}{\sqrt{A_i^2+K\|B_i\|^2}}.
\end{align}
Thus the common factor $Z_i^{-1}$ cancels from the numerator and denominator. The Euclidean linear-attention denominator $Z_i$ is replaced by the curvature-aware normalization $\sqrt{A_i^2+K\|B_i\|^2}$ induced by the \PV{} midpoint. As $K\to0^-$, $\rho_j=\sqrt{1-K\|v_j\|^2}\to1$, so $A_i\to Z_i$ and the expression becomes the Euclidean linear attention output
\begin{equation}
\frac{\sum_j\lambda_{ij}v_j}{\sum_j\lambda_{ij}}.
\end{equation}

\subsection{Euclidean inner product from the Busemann definition}

For completeness, we recall the identity used to motivate Busemann attention. Let $u\in\mathbb{R}^d$ with $\|u\|=1$ and define
\begin{equation}
B^u(x)=\lim_{t\to\infty}(\|x-tu\|-t).
\end{equation}
Then
\begin{equation}
\|x-tu\|-t
=\frac{\|x-tu\|^2-t^2}{\|x-tu\|+t}
=\frac{\|x\|^2-2t\langle x,u\rangle}{\|x-tu\|+t}.
\end{equation}
Since $\|x-tu\|/t\to1$, the denominator divided by $t$ tends to $2$, and therefore
\begin{equation}
B^u(x)=-\langle x,u\rangle.
\end{equation}
For any nonzero $y$, setting $u=y/\|y\|$ gives
\begin{equation}
\langle x,y\rangle=-\|y\|B^{y/\|y\|}(x),
\end{equation}
which is the Euclidean identity generalized by \PV{} Busemann attention.

\section{Implementation and Experimental Details}

\subsection{Transformer block implementation}

The implemented \PVFormer{} block keeps the three attention primitives separated. For full attention, the query and value branches use the homomorphic PV maps in \Cref{subsec:pv_homomorphism}; the key branch uses the point-to-hyperplane map in \Cref{app:pv_key_transform}; the score is the PV Busemann logit in \Cref{eq:pv_busemann_score}; and values are aggregated by the PV midpoint in \Cref{eq:pv_midpoint_closed}. For large graphs, the block switches to the PV linear attention branch in \Cref{subsec:pv_linear_attention}. This branch uses point-valued PV queries, keys, and values and avoids materializing the $N\times N$ Busemann score matrix.

For numerical stability, implementation uses a small $\varepsilon$ in key normalization and logarithm arguments, as shown in \Cref{eq:pv_busemann_score}. Orthogonal maps are parameterized through skew-symmetric generators so that the resulting matrices remain in $SO(d_h)$. This preserves the geometric interpretation of the homomorphic $Q/V$ updates while keeping all trainable parameters unconstrained in Euclidean memory.

\subsection{Graph, text, and vision protocols}

The experiments follow the Hypformer-style protocol used in the main text. Small and medium graph experiments report F1-score on Disease and accuracy on Airport, Cora, Citeseer, and PubMed. Large graph experiments report ROC-AUC on ogbn-proteins and accuracy on the large-scale accuracy benchmarks shown in \Cref{tab:graph}. Text and vision experiments evaluate Mini-ImageNet and 20News-Groups under $k$-nearest-neighbor graphs with $k\in\{5,10,15,20\}$ and a graph-free variant.

The ablations in \Cref{tab:pvnn_fc_qv_ablation,tab:similarity_key_ablation} are designed to isolate geometric components. The PVNN-FC ablation replaces only the homomorphic query/value maps. The similarity/key ablation keeps the graph branch, optimizer, curvature, dropout, homomorphic $Q/V$ maps, and midpoint aggregation fixed while changing either the similarity function or the key map. The attention scalability experiment in \Cref{fig:pv_attention_scalability} uses synthetic attention states with hidden dimension 128 and one head, and reports forward/backward time per attention step and peak GPU memory.

\subsection{Reproducibility notes}

The reported tables use the values in the main manuscript as the experimental record. For small and medium graph ablations, the explicit ablation seeds are five different ones when the table reports a five-seed mean and standard deviation. Large-graph experiments use the scalable PV linear attention path, because full Busemann attention is quadratic in the number of nodes and is not feasible at the large-graph scale shown in \Cref{fig:pv_attention_scalability}. Absolute local paths, machine-specific process identifiers, and private infrastructure details are omitted from the manuscript to keep the supplementary material portable and anonymous. All experiments were conducted on A100 80GB GPUs.

\section{Numerical Stability of \PV{}}

\Cref{tab:app_pv_model_stability,tab:app_pv_attention_stability,tab:app_pv_training_stability} summarize stability at the model, attention-block, and training levels. The model-level results from \PVNN{}~\citep{chen2026proper} show no reported \PV{} failures across the tested radii, with constraint violations inapplicable to its unconstrained representation. Our attention and training diagnostics use FP32 and the official stabilized \Hypformer{} baseline. The attention probe feeds both models the same 512 embeddings at each radius. The Lorentz baseline retains its native safeguards, including time-coordinate reconstruction and argument clamping. \PVFormer{} retains nonzero input gradients with a spread no greater than $1.20$, whereas the Lorentz baseline has $67.6\%$ to $88.9\%$ exactly zero gradients for $r\ge5$. Increasing $k=|K|$ from $0.5$ to $64$ reduces the Lorentz singularity margins by approximately $72\times$ on Airport and $93\times$ on Citeseer, while the \PV{} margins remain comparatively stable. The fixed-radius probe measures robustness under synthetic stress, and no Lorentz training run diverged in the curvature sweep.

\begin{table}[!htbp]
\centering
\caption{Model-level failure and manifold-constraint violation rates (\%), reproduced from \PVNN{}~\citep{chen2026proper}. N/A denotes an inapplicable constraint check. These rates do not measure boundary gradient saturation.}
\label{tab:app_pv_model_stability}
\small
\setlength{\tabcolsep}{2.8pt}
\renewcommand{\arraystretch}{1.1}
\begin{tabular}{@{}ll*{10}{r}@{}}
\toprule
Metric (\%) & Model & \multicolumn{10}{c}{Geodesic radius $r$} \\
\cmidrule(lr){3-12}
& & $1$ & $5$ & $10$ & $20$ & $50$ & $75$ & $100$ & $150$ & $200$ & $1000$ \\
\midrule
Failure rate & \PV{} & $0$ & $0$ & $0$ & $0$ & $0$ & $0$ & $0$ & $0$ & $0$ & $0$ \\
Failure rate & Poincar\'e & $0$ & $0$ & $0$ & $0$ & $0$ & $0$ & $0$ & $0$ & $0$ & $0$ \\
Failure rate & Hyperboloid & $0$ & $0$ & $0$ & $4.23$ & $64.42$ & $79.63$ & $88.26$ & $96.43$ & $100$ & $100$ \\
\midrule
Violation rate & \PV{} & N/A & N/A & N/A & N/A & N/A & N/A & N/A & N/A & N/A & N/A \\
Violation rate & Poincar\'e & $0$ & $0$ & $0$ & $0$ & $0$ & $0$ & $0$ & $0$ & $0$ & $0$ \\
Violation rate & Hyperboloid & $32.50$ & $92.36$ & $99.76$ & $100$ & $100$ & $100$ & $100$ & $100$ & $100$ & $100$ \\
\bottomrule
\end{tabular}
\end{table}

\begin{table}[!htbp]
\centering
\caption{Attention-block stability with 512 embeddings at each radius. Max/min is the gradient spread over nonzero input gradients, and zero-grad is the percentage of exactly zero input gradients. Lorentz denotes the official stabilized \Hypformer{} implementation.}
\label{tab:app_pv_attention_stability}
\small
\setlength{\tabcolsep}{4pt}
\renewcommand{\arraystretch}{1.1}
\begin{tabular}{@{}rcccc@{}}
\toprule
Input radius $r$ & \PV{} max/min & \PV{} zero-grad & Lorentz max/min & Lorentz zero-grad \\
\midrule
$0.5$ & $1.01$ & $0\%$ & $3.35$ & $0\%$ \\
$1$ & $1.03$ & $0\%$ & $23.9$ & $0\%$ \\
$2$ & $1.06$ & $0\%$ & $1.2\times10^{3}$ & $0\%$ \\
$5$ & $1.06$ & $0\%$ & $1.7\times10^{28}$ & $67.6\%$ \\
$10$ & $1.12$ & $0\%$ & $1.9\times10^{10}$ & $88.6\%$ \\
$50$ & $1.20$ & $0\%$ & $2.0\times10^{9}$ & $88.9\%$ \\
\bottomrule
\end{tabular}
\end{table}

\begin{table}[!htbp]
\centering
\caption{Minimum distance of the Lorentz arcosh argument or the \PV{} Busemann logarithm argument from its singularity during training as $k=|K|$ varies. The collapse row summarizes the change from $k=0.5$ to $k=64$. Margins are interpreted within each geometry.}
\label{tab:app_pv_training_stability}
\small
\setlength{\tabcolsep}{5pt}
\renewcommand{\arraystretch}{1.1}
\begin{tabular}{@{}lcccc@{}}
\toprule
$k=|K|$ & Airport: Lorentz & Airport: \PV{} & Citeseer: Lorentz & Citeseer: \PV{} \\
\midrule
$0.5$ & $2.466$ & $0.526$ & $0.574$ & $0.952$ \\
$1$ & $1.530$ & $0.462$ & $0.350$ & $0.939$ \\
$4$ & $0.476$ & $0.642$ & $0.098$ & $0.886$ \\
$16$ & $0.125$ & $0.570$ & $0.024$ & $0.722$ \\
$64$ & $0.0344$ & $0.547$ & $0.00614$ & $0.763$ \\
\midrule
Collapse & $72\times$ & flat & $93\times$ & $1.25\times$ \\
\bottomrule
\end{tabular}
\end{table}

\section{Additional Experiments}

\subsection{Efficiency on real large-scale graphs}
\label{app:real_graph_efficiency}

We extend the attention efficiency comparison in \Cref{fig:pv_attention_scalability} to real large-scale graphs. \Cref{tab:app_real_graph_efficiency} reports the attention-branch time per step on ogbn-arxiv with the full graph and on Amazon2M with two batch sizes. Compared with \Hypformer{}, \PVFormer{} reduces this time by $33.9\%$ on ogbn-arxiv and by $29.7\%$ and $11.7\%$ on Amazon2M with batches of 900K and 100K nodes, respectively.

\begin{table}[!htbp]
\centering
\caption{Attention-branch time per step (ms) on real large-scale graphs. Time reduction is measured relative to \Hypformer{}.}
\label{tab:app_real_graph_efficiency}
\small
\setlength{\tabcolsep}{6pt}
\renewcommand{\arraystretch}{1.1}
\begin{tabular}{@{}lccc@{}}
\toprule
Method & \shortstack{ogbn-arxiv\\(full graph, 169K)} & \shortstack{Amazon2M\\(batch 900K)} & \shortstack{Amazon2M\\(batch 100K)} \\
\midrule
\Hypformer{} & $204.1\pm1.2$ & $929.8\pm2.2$ & $89.5\pm0.2$ \\
\PVFormer{} & $135.0\pm0.3$ & $653.6\pm2.2$ & $79.0\pm0.3$ \\
\midrule
Time reduction & \textbf{33.9\%} & \textbf{29.7\%} & \textbf{11.7\%} \\
\bottomrule
\end{tabular}
\end{table}

During Cora training, \Hypformer{} performs an average of $12.5$ constraint-restoration operations per forward pass, whereas \PVFormer{} requires none because \PV{} coordinates are unconstrained. This count includes time-coordinate reconstruction, hyperboloid projection, and midpoint renormalization. Tangent-space maps themselves are excluded, while projections invoked inside them are counted. The fractional average reflects dropout being active only during training.

\subsection{Gradient stability compared with \PVNN{}}
\label{app:pvnn_gradient_ablation}

\Cref{tab:app_pvnn_gradient_ablation} reports the within-batch gradient spread under different Q/V transformations and similarity functions in the synthetic radius probes. At $r=50$, replacing our homomorphic Q/V map with PVNN-FC while retaining Busemann similarity increases the spread from $1.20$ to $1.14\times10^{3}$. With PVNN-FC fixed, replacing Busemann similarity with distance increases it further to $1.71\times10^{3}$. With homomorphic Q/V maps, the two similarities yield spreads of $1.20$ and $1.25$, respectively. The homomorphic maps preserve norms and constrain expressiveness, so these results characterize gradient balance for the tested designs and should be interpreted together with the expressiveness discussion in \Cref{app:homomorphic_discussion}.

\begin{table}[!htbp]
\centering
\caption{Within-batch gradient spread of the \PV{} attention block under different Q/V transformations and similarity functions.}
\label{tab:app_pvnn_gradient_ablation}
\small
\setlength{\tabcolsep}{4pt}
\renewcommand{\arraystretch}{1.1}
\begin{tabular}{@{}rcccc@{}}
\toprule
Input radius & \shortstack{Homomorphic (ours)\\+ Busemann (ours)} & \shortstack{PVNN-FC\\+ Busemann (ours)} & \shortstack{Homomorphic (ours)\\+ distance} & \shortstack{PVNN-FC\\+ distance} \\
\midrule
$5$ & $1.07$ & $12.8$ & $1.15$ & $18.0$ \\
$10$ & $1.13$ & $551$ & $1.20$ & $726$ \\
$50$ & $1.20$ & $1.14\times10^{3}$ & $1.25$ & $1.71\times10^{3}$ \\
\bottomrule
\end{tabular}
\end{table}

\section{Discussion of Our Homomorphic Transformations}
\label{app:homomorphic_discussion}

\textbf{Current maximality and the choice of $SO(d)$.} In Euclidean space, linear homomorphisms admit the general and tractable form $x\mapsto Ax$. Our construction identifies \PV{} gyro-homomorphisms through the sufficient condition of being an origin-preserving isometry. For equal-dimensional \PV{} transformations, the maximal family satisfying this condition is the orthogonal group $\{x\mapsto Rx\mid R\in O(d)\}$. This maximality concerns origin-preserving isometries, while a classification of more general \PV{} gyro-homomorphisms is outside the scope of this work. Theorem~3.1 of Helfrich et al.~\citep{helfrich2018orthogonal} provides a scaled Cayley parameterization of $O(d)$ through an additional diagonal sign matrix $D$, with the number $\rho$ of its negative entries treated as a hyperparameter. For simpler optimization, we use $SO(d)$.

\textbf{Expressiveness, dimension-changing cases, and future work.} Our sufficient-condition construction identifies a tractable subclass of \PV{} gyro-homomorphisms. Its rotation constraint preserves norms and may limit expressiveness relative to more general transformations. A dimension-reducing map with $d_k<d$ is not covered by our current homomorphic construction, since the present work considers only equal-dimensional transformations. Characterizing broader \PV{} gyro-homomorphisms and developing dimension-changing homomorphic maps are two important directions for future work.

\section{Limitations}
\label{app:limitations}

Due to limited computational resources, we have not trained \PVFormer{} at the scale of large language models. We plan to apply it to large language models in future work. Our homomorphic Q/V maps are restricted to equal-dimensional rotations, with expressiveness and dimension-changing extensions discussed in \Cref{app:homomorphic_discussion}. The numerical stability diagnostics characterize the tested operators and stress regimes.